\documentclass[mnsc,nonblindrev]{informs3} 
\usepackage{wrapfig,lipsum,booktabs}

\OneAndAHalfSpacedXI 

\usepackage{natbib}
 \bibpunct[, ]{(}{)}{,}{a}{}{,}%
 \def\bibfont{\small}%
\usepackage{enumitem}
\usepackage{float}

\usepackage{amsmath,amsfonts,amssymb}
\newtheorem{asu}{Assumption}
\newcounter{subassumption}[asu]

\DeclareRobustCommand{\Qipw}{Q^{\text{IPW}}}
\DeclareRobustCommand{\Qdm}{Q^{\text{DM}}}
\DeclareRobustCommand{\Qdr}{Q^{\text{DR}}}

\DeclareRobustCommand{\QipwI}{Q^{\text{IPW}}_I}
\DeclareRobustCommand{\QdmI}{Q^{\text{DM}}_I}
\DeclareRobustCommand{\QdrI}{Q^{\text{DR}}_I}
\DeclareRobustCommand{\QkptI}{Q^{\text{K-PT}}_I}
\DeclareRobustCommand{\QbptI}{Q^{\text{B-PT}}_I}

\DeclareRobustCommand{\QipwN}{Q^{\text{IPW}}_N}
\DeclareRobustCommand{\QipwIN}{Q^{\text{IPW}}_{N,I}}
\DeclareRobustCommand{\muhatN}{\hat \mu^{N}}

\DeclareRobustCommand{\QdrN}{Q^{\text{DR}}_N}
\DeclareRobustCommand{\QdrIN}{Q^{\text{DR}}_{N,I}}
\DeclareRobustCommand{\nuhatN}{\hat \nu^{N}}

\DeclareRobustCommand{\QdmIN}{Q^{\text{DM}}_{N,I}}

\DeclareRobustCommand{\epsIN}{\epsilon_{N,I}}

\DeclareRobustCommand{\vhatINipw}{\hat v^{\text{IPW}}_{N,I}}
\DeclareRobustCommand{\shatINipw}{\hat{\sets S}^{\text{IPW}}_{N,I}}

\DeclareRobustCommand{\vhatINdr}{\hat v^{\text{DR}}_{N,I}}
\DeclareRobustCommand{\shatINdr}{\hat {\sets S}^{\text{DR}}_{N,I}}

\DeclareRobustCommand{\vhatINdm}{\hat v^{\text{DM}}_{N,I}}
\DeclareRobustCommand{\shatINdm}{\hat {\sets S}^{\text{DM}}_{N,I}}

\DeclareRobustCommand{\vstar}{ v^{\star}}
\DeclareRobustCommand{\sstar}{{\sets S}^{\star}}

\makeatletter
\renewcommand{\p@subassumption}{\theasu}
\makeatother

\newcommand\independent{\protect\mathpalette{\protect\independenT}{\perp}}
\def\independenT#1#2{\mathrel{\rlap{$#1#2$}\mkern2mu{#1#2}}}

\usepackage{bbm}
\usepackage{graphicx}
\usepackage{enumitem}
\graphicspath{ {./images/} }

\TheoremsNumberedThrough     

\EquationsNumberedThrough    

\newcommand{\field}[1]{\ensuremath{\mathbb{#1}}}
\newcommand{\sets}[1]{\ensuremath{\mathcal{#1}}}

\newcommand{\reals}{\ensuremath{\field{R}}} 

\usepackage[dvipsnames,table,xcdraw]{xcolor}
\usepackage{natbib}
\usepackage{hyperref}

\newcommand{\revision}[1]{{#1}}
\newcommand{\revisio}[1]{{#1}}

\newcommand{\newpv}[1]{{{#1}}}
\newcommand{\todiscusspv}[1]{{\color{red} [{\sffamily\bfseries TO DISCUSS PV:} {\em #1}]}}

\newcommand{\comment}[1]{}

\newcommand{\delag}[1]{}

\newcommand{\newsa}[1]{{#1}}

\newcommand{\newnj}[1]{{#1}}

\MANUSCRIPTNO{}
\renewcommand{\theARTICLETOP}{}

\begin{document}



\RUNAUTHOR{Jo et al.}
\RUNTITLE{Learning Optimal Prescriptive Trees from Observational Data}

\TITLE{Learning Optimal Prescriptive Trees \\ from Observational Data}

\ARTICLEAUTHORS{%
\AUTHOR{Nathanael Jo}
\AFF{University of Southern California, Los Angeles, CA 90089, \EMAIL{nathanael.jo@gmail.com}} 


\AUTHOR{Sina Aghaei}
\AFF{University of Southern California, Los Angeles, CA 90089, \EMAIL{saghaei@usc.edu}}

\AUTHOR{Andrés Gómez}
\AFF{University of Southern California, Los Angeles, CA 90089, \EMAIL{gomezand@usc.edu}}

\AUTHOR{Phebe Vayanos}
\AFF{University of Southern California, Los Angeles, CA 90089, \EMAIL{phebe.vayanos@usc.edu}}
} 

\ABSTRACT{%
We consider the problem of learning an optimal prescriptive tree (i.e., an interpretable treatment assignment policy in the form of a binary tree) of moderate depth, from observational data. This problem arises in numerous socially important domains such as public health and personalized medicine, where interpretable and data-driven interventions are sought based on data gathered in deployment -- through passive collection of data -- rather than from randomized trials. We propose a method for learning optimal prescriptive trees using mixed-integer optimization (MIO) technology. We show that under mild conditions our method is asymptotically exact in the sense that it converges to an optimal out-of-sample treatment assignment policy as the number of historical data samples tends to infinity. \revisio{Contrary to existing literature, our approach: \textit{1)}~does not require data to be randomized, \textit{2)}~does not impose stringent assumptions on the learned trees, and \textit{3)}~has the ability to model domain specific constraints. Through extensive computational experiments, we demonstrate that our asymptotic guarantees translate to significant performance improvements in finite samples, as well as showcase our uniquely flexible modeling power by incorporating budget and fairness constraints.}
}%



\maketitle


\section{Introduction}
\label{section:intro}

\newnj{Prescriptive analytics} is concerned with determining the best treatment for an individual based on their personal characteristics. This problem arises in a variety of domains, from optimizing online advertisements for different users \citep{li2010contextual} to assigning suitable therapies for patients suffering from a medical condition \citep{flume2007cystic}. In this paper, we aim to design such personalized policies based on data collected in deployment rather than during randomized trials. In such settings, historical treatments are not assigned at random but rather based on a (potentially unknown) policy (e.g., social workers or doctors triaging clients/patients based on their personal characteristics or symptoms). Being able to design personalized treatment assignment policies without running randomized trials is extremely important since such trials are often considered unethical, impractical, or may take too long to conduct.

\newnj{We are particularly motivated by decision- and policy-making in high-stakes domains \revision{-- for example, to assign scarce housing resources to those experiencing homelessness~\citep{azizi2018designing}; or to match patients to scarce organs for transplantation~\citep{bertsimas2013fairness}}. In such contexts (contrary to, say, online advertising), policies are designed \emph{offline} and need to be \emph{(a)} highly \emph{interpretable}, so that system stakeholders (e.g., policy-makers and populations on which the system is deployed) can audit and scrutinize the decision process; and \emph{(b)} \emph{optimal}, so as to optimize outcomes on the population among models in a given class. We take the view that one should opt for simpler models that are themselves interpretable rather than learning black-box models and then explaining their decisions -- the latter of which has a burgeoning literature, see e.g.,~\cite{linardatos2020explainable}. There are, admittedly, trade-offs between designing and learning interpretable models that maximize performance versus justifying a black-box model's decisions ex-post; we refer the interested reader to \cite{rudin2019stop} for a thorough discussion.}

\newnj{With these needs in mind, we focus our attention on the design of \emph{optimal prescriptive trees}}. A prescriptive tree takes the form of a binary tree. In each branching node of the tree, a binary test is performed on a feature. Two branches emanate from each branching node, with each branch representing the outcome of the test. If a datapoint passes (resp.\ fails) the test, it is directed to the left (resp.\ right) branch. A treatment is assigned to all leaf nodes. Thus, each path from root to leaf represents a treatment assignment rule that assigns the same treatment to all datapoints that reach that leaf. \newnj{Similar to decision trees in the context of machine learning~\citep{rudin2019stop}, prescriptive trees are some of the most interpretable models. \revision{We also note that we focus on binary trees in line with most works on decision trees (e.g., the C4.5 and CART~\citep{breiman2017classification} algorithms). In general, non-binary trees can be modeled using binary trees with more branching decisions, and there is increasingly work being done on non-binary trees (see e.g.,~\cite{mazumder2022quant}).}}

The goal of learning prescriptive trees is to select branching decisions and treatment assignments such that the expected value of the outcomes of the treatments on the population are maximized. We say that a prescriptive tree is \emph{optimal} (in-sample, given a training set) if there exists a mathematical proof that no other tree yields better expected outcomes in the population used for training the method. Accordingly, we will say that a tree is \emph{optimal out-of-sample} if there exists a proof that no other tree yields better expected outcomes in the entire population.


\subsection{Problem Statement}
\label{section:problem_statement}

We now formalize the problem we study. In our discussion, we use terminology and concepts from \cite{hernan2019causal}. The goal is to design a personalized policy $\pi : \sets X \rightarrow \sets K$ in the form of a prescriptive tree that maps an individual's characteristics $x \in \sets X \subseteq \reals^F$ to a treatment from a finite set of candidates indexed in the set $\sets{K}$. We note that the set $\sets K$ may also include a ``no treatment'' option. Each individual is characterized by their covariates $X \in \reals^F$ and their \emph{potential outcomes} $Y(k) \in \sets Y \subseteq \reals$ under each treatment $k \in \sets K$\comment{, see e.g., \cite{hernan2019causal}}. The joint distribution of $X$ and $\{Y(k)\}_{k \in \sets K}$ is unknown. However, we have access to $I$ \newnj{i.i.d.} historical observations $\sets D:= \{(X_i, K_i, Y_i)\}_{i \in \mathcal{I}}$ indexed in the set $\sets I :=\{1,\ldots, I\}$, where $X_i$ denote the covariates of the $i$th observation, $K_i$ is the treatment assigned to it, and $Y_i = Y_i(K_i)$ is the \emph{observed outcome}, i.e., the outcome under the treatment received\comment{, see e.g., \cite{hernan2019causal}}. We use the convention that higher values of $Y_i$ are more desirable. Critically, as in all observational experiments, we cannot control the historical treatment assignment policy and the outcomes $Y_i(k)$ for $k \neq K_i$ remain \emph{unobserved}. Formally,  our aim is to learn, from the observational data $\sets D$, a policy \revisio{$\pi \in \Pi_d$} that maximizes the quantity
$$
Q(\pi) \; := \;  \mathbb{E}\left[Y(\pi(X))\right],
$$
where \revisio{$\Pi_d$ denotes the set of prescriptive trees of maximum depth $d$, and} $\mathbb E(\cdot)$ denotes the expectation operator with respect to the joint distribution $\mathbb P$ of $X, K, Y(1), \ldots, Y(k)$. A major challenge in learning the best policy $\pi$ is the fact that we cannot observe the \emph{counterfactual} outcomes $Y_i(k)$, $k\in \sets K, \; k \neq K_i$ that were not received by datapoint~$i$.  Indeed, the missing data make it difficult to identify the best possible treatment for each datapoint. In fact, without further assumptions on the historical policy, it is impossible to even identify the treatment that is the best \emph{in expectation} for any given datapoint.


\subsection{Background on Causal Inference}\label{sec:background_ci}

In this section, we provide a brief overview of the tools from causal inference that motivate the assumptions we make in our work and our proposed methods.

\subsubsection{Identifiability, Randomized Experiments, and Observational Studies.}
\label{section:randomization}

Learning an optimal prescriptive tree from the observed data requires that the expected values of the counterfactual outcomes be \emph{identifiable}, i.e., possible to be expressed as a function of the observed data. A sufficient condition for identifiability is for the data to have been collected during a \emph{(marginally) randomized} or \emph{conditionally randomized} experiment, see \cite{hernan2019causal}.

In a randomized experiment, \comment{treatments are assigned at random and }all individuals have the same chance of getting any given treatment independent of their characteristics $x$. \comment{(although treatments need not be equally probable)} Formally, we have $K \independent X$. When treatment assignment is randomized, we say that treatment groups are \emph{exchangeable} because it is irrelevant which group received a particular treatment. Exchangeability means that for all $k$ and~$k' \in \sets K$, those who got treatment~$k$ in the data would have the same outcome distribution under $k'$ as those who got treatment~$k'$. Equivalently, exchangeability means that the counterfactual outcome and the actual treatment are independent, or~$Y(k) \independent K$, for all values of~$k$.

In a conditionally randomized experiment, treatment assignment probabilities depend on the characteristics $x$ of each individual. A conditionally randomized experiment will, in general, not result in exchangeability because treatment groups may differ systemically from one another. However, treatment groups are exchangeable \emph{conditionally on~$X$} since no information other than~$X$ is used to decide the treatment. Thus, conditional randomization ensures \emph{conditional exchangeability}, i.e., $Y(k) \independent K \; | \; X$, for all values of~$k$. We can then infer the outcome distribution under treatment $k'$ for those with covariates $x$ that got treatment $k$ in the data by looking at their counterparts with the same (or similar) $x$ that received treatment $k'$.

As mentioned previously, in high-stakes domains we typically only have access to data from an \emph{observational study}. In such contexts, it is common practice in the literature on causal inference to assume that treatment is assigned at random conditional on $X$, although this may be an approximation and not possible to check in practice. The observational study can then be viewed as a conditionally randomized experiment, provided: \emph{a)} interventions are well defined and are recorded as distinct treatment values in the data; \emph{b)} the conditional probability of receiving a treatment depends only on the covariates $X$; and \emph{c)} the probability of receiving any given treatment conditional on $X$ is positive, i.e., $\mathbb P(K=k|X=x)>0$ almost surely for all $k$. This last condition is termed \emph{positivity} and holds in practice for randomized and conditionally randomized experiments.

We now discuss several methods from the literature for evaluating the performance of a \emph{counterfactual} treatment assignment policy $\pi$ from (conditionally) randomized experiments or, more generally, from observational studies that satisfy the above conditions. In particular, in the following and throughout the paper, we assume that treatments in the data have been assigned according to a (potentially unknown) \emph{logging policy} $\mu$, where $\mu(k,x) := \mathbb P(K=k|X=x)$ and $\mu(k,x)>0$ for all $k \in \sets K$ and $x \in \sets X$. 


\subsubsection{Inverse Propensity Weighting\newpv{.}}
\label{section:ipw_intro}

The performance of a counterfactual treatment assignment policy $\pi$ can be evaluated using Inverse Probability Weighting (IPW). 

IPW was originally proposed to estimate causal quantities such as expected values of counterfactual outcomes, average treatment effects, risk ratios, etc., see \cite{horvitz1952generalization}. It relies on reweighting the outcome for each individual $i$ in the dataset by the inverse of their \emph{propensity score}, given by $\mu(K_i,X_i)$. Such reweighting has the effect of creating a \emph{pseudo-population} where all individuals in the data are hypothetically given all treatments (thus simulating a randomized treatment assignment). Doing so allows for the estimation of the distribution of the unobserved counterfactual outcomes for all possible values of $X$.

\begin{figure}[h]
\includegraphics[width=\textwidth]{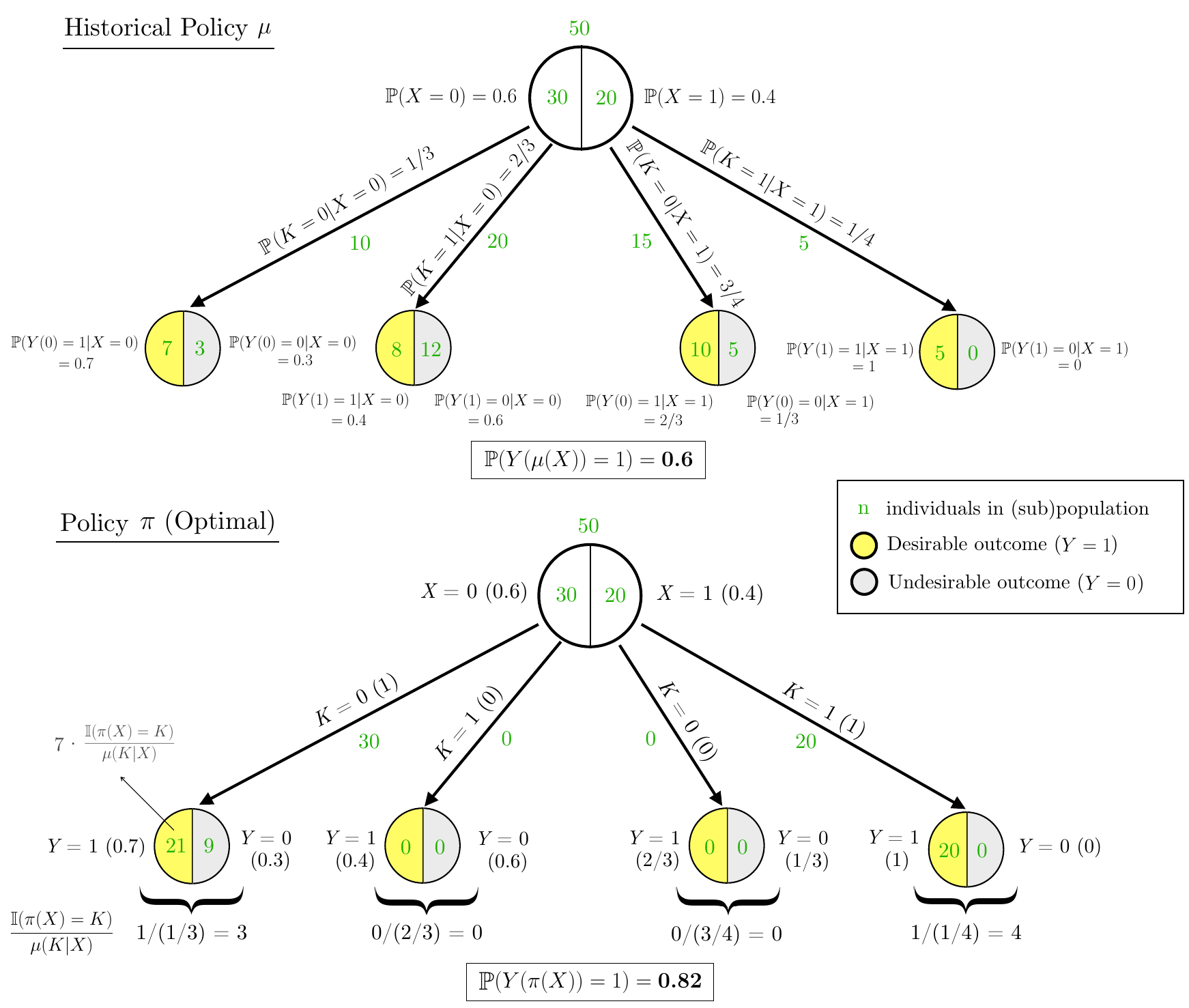}
\centering
\caption{Figure illustrating the evaluation of the performance of the counterfactual policy~$\pi$ (bottom tree) from data collected using the logging policy~$\mu$ (top tree) using the IPW estimator--notational conventions based on~\cite{hernan2019causal}. The historical data presents 50 instances, each with a single binary covariate $X^1 \in \{0,1\}$. $X^1 = 1$ (resp.\ $0$) indicates that the patient is sick (resp. healthy). There are two treatment options ($\sets K:=\{0,1\}$), where $K=1$ ($0$) is to (not) treat the patient. The potential outcomes are binary ($Y(k) \in \{0,1\}$, $k \in \sets K$). The expected reward under policy $\mu$ is $0.6$, while policy $\pi$ has a $0.82$ probability of getting a positive outcome--in fact, it can be shown that $\pi$ is optimal.}
\label{fig:ipw_policy_example}
\end{figure}

Importantly, IPW can also be used to estimate the performance of a counterfactual policy~$\pi$ by reweighting each individual~$i$ in the data by $\frac{\mathbbm{1}(\pi(X_i) = K_i)}{\mu(K_i, X_i)}$ as illustrated in Figure~\ref{fig:ipw_policy_example}, see also \cite{charles2013counterfactual}. This reweigthing simulates the performance of the policy $\pi$ by leveraging the conditional exchangeability property of the data. Using this estimator, the quantity $Q(\pi)$ is estimated from the data as
\begin{equation}\label{eq:ipw_obj}
\QipwI(\pi) := \frac{1}{I}\sum_{i \in \mathcal I} \frac{\mathbbm{1}(\pi(X_i) = K_i) }{\hat \mu(K_i, X_i)} Y_i,
\end{equation}
where $\hat \mu$ is an estimator of $\mu$, which is obtained for instance using machine learning, by fitting a model to $\{(X_i,K_i)\}_{i \in \sets I}$. If $\mu$ is known or if~$\hat \mu$ converges almost surely to~$\mu$, the IPW estimator of $Q(\pi)$ is \newnj{statistically consistent}. However, it may suffer from high variance, particularly if some of the propensity scores $\mu(K, X)$ are small, see~\cite{Dudik2011DoublyLearning}.

\subsubsection{Direct Method.} \label{section:dm}

An alternative estimator of the performance of the counterfactual policy $\pi$ is the Direct Method (DM)\comment{, see \cite{Dudik2011DoublyLearning}}.

DM is based upon the so-called Regress and Compare (R\&C) approach, which proceeds in three steps to design a policy that maximizes the expected value of the outcomes \citep{Kallus2018BalancedLearning}. First, it partitions the dataset $\sets{D}$ by treatments. Second, for all $k\in \sets K$, it learns a model $\hat{\nu}_{k}(X)$ of $\mathbb{E}(Y|K=k, X)$ using the subpopulation that was assigned treatment $k$. Third, given any $X\in \reals^F$, it estimates $Y(k)$ as $\hat{\nu}_{k}(X)$ and assigns the best treatment according to $\hat \nu$, that is, $\pi^{\text{R\&C}}(X) = \argmax_{k \in \sets{K}} \hat{\nu}_{k}(X)$. R\&C has been used, for example in the online setting by \cite{bastani2020online} and \cite{qian2011performance}. Usually, variants of linear regression are used to get estimators $\hat{\nu}$, see \cite{bastani2020online,goldenshluger2013linear,li2010contextual}.

The R\&C approach, however, does not allow the evaluation of arbitrary policies nor does it enable the design of policies that belong to an arbitrary class. DM adapts the regress and compare framework to resolve these shortcomings. It proposes to estimate $Q(\pi)$ via
\begin{equation}\label{eq:dm_obj}
\QdmI(\pi) := \frac{1}{I}\sum_{i \in \mathcal I} \hat{\nu}_{\pi(X_i)}(X_i).
\end{equation}
Both R\&C and DM result in poor performance if $\hat{\nu}_k(X)$, $k\in \sets K$, are biased estimators or make predictions that are far from the true expected outcomes, see~\cite{beygelzimer2009offset}. As pointed out by~\cite{Dudik2011DoublyLearning}, $\hat{\nu}_k(X)$ are constructed without information about~$\pi$ and thus may approximate $\mathbb E(Y|K=k,X)$ poorly in areas that are relevant for $Q(\pi)$. That said, if $\hat \nu_k$ are \newnj{statistically consistent} for all $k \in \sets K$, the DM estimator of~$Q(\pi)$ will also be \newnj{consistent}.


\subsubsection{Doubly Robust Approach.}
\label{section:robust_intro}

Doubly robust (DR) estimation is a family of techniques that combine two estimators. If either one of the two estimators is accurate, then the doubly robust approach is also accurate, which dampens the errors brought by the individual estimators.

In particular, realizing the drawbacks of IPW and DM, \cite{Dudik2011DoublyLearning} proposed a doubly robust approach that estimates the counterfactual performance of a policy $\pi$ as
\begin{equation}
\label{eq:dr_obj}
\QdrI(\pi) := \frac{1}{I}\sum_{i \in \mathcal I} \left(\hat{\nu}_{\pi(X_i)}(X_i)+ (Y_i - \hat{\nu}_{K_i}(X_i)) \frac{\mathbbm{1}(\pi(X_i) = K_i)}{\hat \mu(K_i, X_i)}\right).
\end{equation}
Intuitively, DR uses IPW to estimate the difference between the true outcome and the prediction made by the DM estimator, $Y_i - \hat{\nu}_{K_i}(X_i)$, thus removing any bias caused by DM, assuming the IPW estimator is \newnj{statistically consistent}. Provided at least one of $\hat \mu$ or $\hat \nu$ converges almost surely to $\mu$ or $\nu$, respectively, the DR estimator of $Q(\pi)$ will be asymptotically \newnj{consistent}. Furthermore, in practice, $\QdrI$ usually has a smaller variance compared to $\QipwI$ but a higher variance than $\QdmI$, see~\cite{Dudik2011DoublyLearning}.


\subsection{Related Works}\label{section:related_works}
In this section, we position our work in the fields of operations research, causal inference, and machine learning.

\subsubsection{Estimating Heterogeneous Causal Effects.}
To estimate heterogeneous causal effects, \cite{Athey2016RecursiveEffects} and \cite{wager2018estimation} use recursive partitioning to produce causal trees and causal forests, respectively. Non-tree-based methods can be found in \cite{abrevaya2015estimating}, which use inverse propensity weighting, and \cite{fan2020estimation}, which extends \cite{abrevaya2015estimating} to account for high-dimensional data. \cite{powers2018some} adapt three methods--random forests, boosting, and MARS (Multivariate Adaptive Regression Splines)--to also estimate treatment effects in high-dimensional data. The overall goal of these approaches is to find good estimates of the treatment effects rather than to design treatment assignment policies. While some of these, such as the method from \cite{Athey2016RecursiveEffects}, can be \emph{adapted} to design tree policies, these policies will not be interpretable due to the need to combine the predictive trees associated with many different treatments.



\subsubsection{Mixed-Integer Optimization.}

MIO has been gaining traction as a tool to solve challenging learning problems. It has been used for example to tackle sparse regression problems \citep{wilson2017alamo, atamturk2019rank,bertsimas2020sparse,hazimeh2020sparse,xie2020scalable,gomez2021mixed}, verification of neural networks \citep{fischetti2018deep,khalil2018combinatorial,tjandraatmadja2020convex}\newpv{,} and sparse principal component analysis \citep{dey2018convex, bertsimas2020solving}, among others. More importantly in the context of this paper, MIO methods have been proposed to learn optimal decision trees \citep{bertsimas2017optimal, verwer2019learning, aghaei2019learning, aghaeistrong, elmachtoub2020decision, mivsic2020optimization}.

MIO technology has also been used in causal inference. Existing approaches have mostly focused on matching \citep{bennett2020building,zubizarreta2017optimal}, i.e., to estimate causal effects by pairing each treatment instance with a control instance that has similar covariates \citep{rubin2006matched}. A related approach to matching that utilizes MIO technology is subset selection to achieve \emph{balance} (i.e., similar distributions) between the covariates in the treatment and control groups \citep{nikolaev2013balance}. Finally, \citet{mintz2017behavioral} use MIO to estimate the effects of weight-loss interventions in myopic agents and design optimal policies from the perspective of a healthcare provider.

\subsubsection{Learning Prescriptive Trees.}
\label{sec:lit_prescriptive_trees}

In a recent work closely related to this paper, \cite{Kallus2017RecursiveData} proposes an MIO formulation to learn optimal prescriptive trees that relies on a critical assumption: that trees are sufficiently deep such that the historical treatment assignment and covariates are independent \emph{at each leaf}. Building upon this approach, \citet{Bertsimas2019OptimalTrees} seek to ameliorate it by augmenting the objective function with a term aimed to improve accuracy of the outcome predictions. The paper uses coordinate descent with multiple starts to train the prescriptive trees (as opposed to MIO technology), and empirically verifies that the resulting solutions are indeed high-quality. We provide an in-depth analysis of both methods in Section~\ref{section:prev_mio} and compare to them in our experiments.



\subsection{Contributions}\label{section:contributions}

In this work, we do an in-depth analysis of existing methods to learn prescriptive trees, and propose new MIO formulations for solving such problems. In particular, our key contributions are:
\begin{enumerate}[label=\emph{(\alph*)}]
    \item
    We analyze the assumption made by the existing methods to learn prescriptive trees discussed in Section~\ref{sec:lit_prescriptive_trees}. We demonstrate, by means of examples, that when the assumption fails to hold \emph{for all feasible trees}, the trees returned as ``optimal'' by these methods may be severely suboptimal. They may even branch on noise covariates that are not predictive of outcomes nor treatment assignments. Moreover, our examples show that this assumption is very strict in the sense that trees deep enough to allow branching on all levels of all covariates (including covariates that are independent of outcomes and treatment assignments) may be needed for this assumption to hold at an optimal tree.
    \item Motivated by the limitations of existing approaches, we propose novel MIO formulations for learning optimal prescriptive trees based on tools from causal inference. Contrary to methods from the literature, our approaches enable the design of shallow and/or deep trees, thus allowing the decision-maker to tune the trade-off between prescription accuracy and interpretability. We demonstrate that, under mild conditions,  our methods are asymptotically exact in the sense that they converge to an optimal out-of-sample treatment assignment policy as the number of historical data samples tends to infinity. Based on extensive computational experiments on both synthetic and real data, we demonstrate that our asymptotic guarantees translate to significant out-of-sample performance improvements even in finite samples.
\end{enumerate}

The remainder of this paper is organized as follows. In Section \ref{section:prev_mio}, we present the existing prescriptive tree formulations from the literature and study their performance in the context of two simple synthetic examples. We introduce our proposed MIO formulations for learning optimal prescriptive tree in Section~\ref{section:proposed_formulations}. Finally, we summarize our computational experiments in Section~\ref{section:experiments}. All proofs and detailed computational results are provided in the Electronic Companion.



\section{Existing Prescriptive Tree Formulations}
\label{section:prev_mio}
This section presents the methods introduced by \cite{Kallus2017RecursiveData} and \cite{Bertsimas2019OptimalTrees} -- from hereon, we refer to these as K-PT and B-PT, respectively. It further analyzes their advantages and drawbacks using two simple examples.

To describe these two approaches, we work with the notation introduced in~\cite{Kallus2017RecursiveData}, which views \newpv{any tree $\pi$} as producing a partition \newpv{$\sets X = \bigcup_{n \in \sets{L}^\pi} \sets{X}_n^\pi$ such that $\sets{X}_n^\pi \cap \sets{X}_m^\pi = \emptyset$ whenever $n \neq m$, where $\sets L^\pi$} collects the set of leaf nodes \newpv{of $\pi$} and \newpv{$\sets X_n^\pi \subseteq \sets X$} represents the set of covariate values that are associated with datapoints landing in leaf \newpv{$n \in \sets L^\pi$}. The approaches K-PT and B-PT rely on the following \emph{crucial} assumption about the structure of \newpv{all feasible} prescriptive tree\newpv{s}.

\begin{asu}[\cite{Kallus2017RecursiveData}]
    \revision{All prescriptive trees $\pi \in \Pi_d$ produce a partition $\{\sets X_n^\pi \}_{n \in \sets L^\pi}$ that is sufficiently fine such that }
    $K \independent X \; | \; X \in \sets{X}_n^\pi \quad \forall n \in \newnj{\sets{L}^\pi}.$
    \label{asu:sufficient_partition}
\end{asu}

\revisio{Assumption~\ref{asu:sufficient_partition} is implicitly made in \cite{Kallus2017RecursiveData} \newpv{when the result in} Corollary~3 is used to obtain the MIO formulation. This assumption }is automatically satisfied in randomized experiments (since $K \independent X$ unconditionally) or in conditionally randomized experiments provided the partition is ``sufficiently fine'' -- for example, \newpv{if all feasible trees are such that} \newpv{$\sets{X}_n^\pi$} is a singleton \newpv{for all $n \in \sets L^\pi$}.


\subsection{K-PT}
\label{section:kallus}

Given a \newpv{policy $\pi$ with partition $\{\sets X_n\}_{n \in \sets{L}}$ satisfying the ``sufficiently fine'' condition in} Assumption~\ref{asu:sufficient_partition}, \cite{Kallus2017RecursiveData} proves that the \emph{observable} quantity $\mathbb E(Y| K=k, X \in \sets X_n)$ is an unbiased estimator of $\mathbb E(Y(k)|X \in \sets X_n)$. In other words, to estimate the potential outcomes under treatment~$k$ at leaf~$n$, it suffices to average outcomes for those that actually received treatment~$k$ at leaf~$n$ under~$\mathbb P$; this quantity can be estimated from the observed data. This statement is intuitive under Assumption~\ref{asu:sufficient_partition}, which implies that all individuals with covariates in the set $\sets X_n$ have the same chance of getting any given treatment, i.e., they are exchangeable within $\sets X_n$. It follows that, under Assumption~\ref{asu:sufficient_partition}, the outcomes of a policy $\pi$ that assigns treatment $\pi(\sets X_n) \in \sets K$ to all individuals that land in leaf~$n$ can be estimated as
\begin{equation}
\sum_{n \in \sets L} \mathbb P ( X \in \sets X_n) \cdot \mathbb E ( Y | K= \pi(\sets X_n), X \in \sets X_n).
\label{eq:kallus_estimator}
\end{equation}
Accordingly, this quantity can be estimated from data as
$$
\QkptI(\pi) := \sum_{n\in \sets{L}} \frac{\sum_{i \in \sets{I}}\mathbbm{1}[X_i \in \sets{X}_n]}{|\sets I|}\widehat{Y}_n \left(\pi(\sets{X}_n)\right),
$$
where $\widehat{Y}_n(k)$ is the sample average approximation of $\mathbb E(Y| K=k, X \in \sets X_n)$, defined through
\begin{equation}
\widehat{Y}_n(k):=\frac{1}{\sum_{i \in \sets{I}}\mathbbm{1}[K_{i} = k, X_{i} \in \sets{X}_n]}\sum_{i\in \sets{I}}\mathbbm{1}[K_i =k, X_i \in \sets{X}_n] Y_i.
\label{eq:avg_estimate}
\end{equation}

\subsubsection*{Analysis of K-PT.}

\cite{Kallus2017RecursiveData} proved that $\QkptI(\pi)$ is an unbiased estimator of $Q(\pi)$ when Assumption~\ref{asu:sufficient_partition} holds, which is the case in \emph{marginally} randomized settings for example. A salient advantage of K-PT is that, if Assumption 1 \emph{does} hold, it avoids predicting~$\mu$ and thus avoids the bias caused by potentially inaccurate estimates~$\hat \mu$. Unfortunately, though, in \emph{conditionally} randomized settings, decision trees will in general not satisfy Assumption~\ref{asu:sufficient_partition} and the tree obtained by maximizing $\QkptI(\pi)$ may be severely suboptimal for the problem of maximizing $Q(\pi)$. In fact, as we now demonstrate, K-PT may actively choose trees that violate Assumption~\ref{asu:sufficient_partition} even when the complete out-of-sample distribution is available \emph{and} the tree is deep enough to allow branching on all covariates that are predictive of outcomes and treatment assignments.

\begin{example}
Consider the following data distribution. Let the covariate vector $X = (X^1, X^2) \in \{0,1\}^2$, where $X^1$ and $X^2$ are independent and Bernoulli distributed with parameter 0.5. We interpret $X^1$ to denote the severity of a patient's condition, where $X^1 = 0$ (resp.\ $X^1=1$) indicates a healthy (resp.\ sick) patient. $X^2$ is a noise variable, unrelated to both the outcomes and the treatment assignments. There are two possible treatments, where $K=1$ (resp.\ $K=0$) indicates that the patient was treated (resp.\ not treated). Treatments are assigned according to a conditionally randomized experiment, with $\mathbb P(K=0|X^1=0) = 0.9$ and $\mathbb P(K=0|X^1=1) = 0.1$, i.e., 90\% of healthy patients do not receive the treatment, and 90\% of sick patients do. \newnj{Thus, the assumptions of conditional exchangeability and positivity hold in this setting, see Section~\ref{section:randomization} for definitions and pointers to the relevant literature. The above numbers imply that $\mathbb{P}(K=0)=\mathbb{P}(K=1)=0.5$ in the population}. Table~\ref{table:expected} shows the expected values of the potential outcomes in dependence of the covariate values. \revision{For simplicity, we assume that: \textit{1)} the potential outcomes in Table~\ref{table:expected} have zero variance, meaning there is no uncertainty in the outcomes conditional on $X$; and \textit{2)}} we have sufficient historical data at our disposal so that we can work with the true values in~\eqref{eq:kallus_estimator} rather than with their empirical estimates provided in the definition of~$\QkptI(\pi)$.

\begin{table}[H]
\centering
 \begin{tabular}{c | c c} 
 \hline
 $X^1$ & $\mathbb{E}(Y(0))$ & $\mathbb{E}(Y(1))$ \\
 \hline\hline
 0 & 1 & 0.8 \\ 
 \hline
 1 & 0 & 0.2 \\
 \hline
\end{tabular}
\caption{Companion table to Example~\ref{ex:example1}.  Expectation of potential outcomes in dependence of the covariate values. A value $X^1 = 0$ (resp.\ $X^1=1$) of the covariate indicates a healthy (resp.\ sick) patient. Larger values of the outcomes are preferred.}
\label{table:expected}
\end{table}

From Table~\ref{table:expected}, it can be seen that the (unique) optimal policy treats sick patients only. In particular, this optimal policy can be modeled by a prescriptive tree of depth one (i.e., a tree with a single branching node and two leafs) that branches on~$X^1$, does not treat any of the healthy patients (left leaf) and treats all sick patients (right leaf). The expected outcome under this policy is given by 
$$
\mathbb{E}(Y(0)|X^1=0)\mathbb{P}(X^1=0)+\mathbb{E}(Y(1)|X^1=1)\mathbb{P}(X^1=1)=1\times\frac{1}{2}+0.2\times \frac{1}{2}=0.6.
$$
One may similarly calculate the expected outcomes under a policy that branches on the noise feature~$X^2$. In this case, no matter which treatment is assigned at each leaf, the resulting policies would have a lower expected outcome of 0.5. These candidate policies and the corresponding expected outcomes under each policy are illustrated in the top row of Figure~\ref{fig:kallus_example}.

\begin{figure}[h]
	\includegraphics[width=\textwidth]{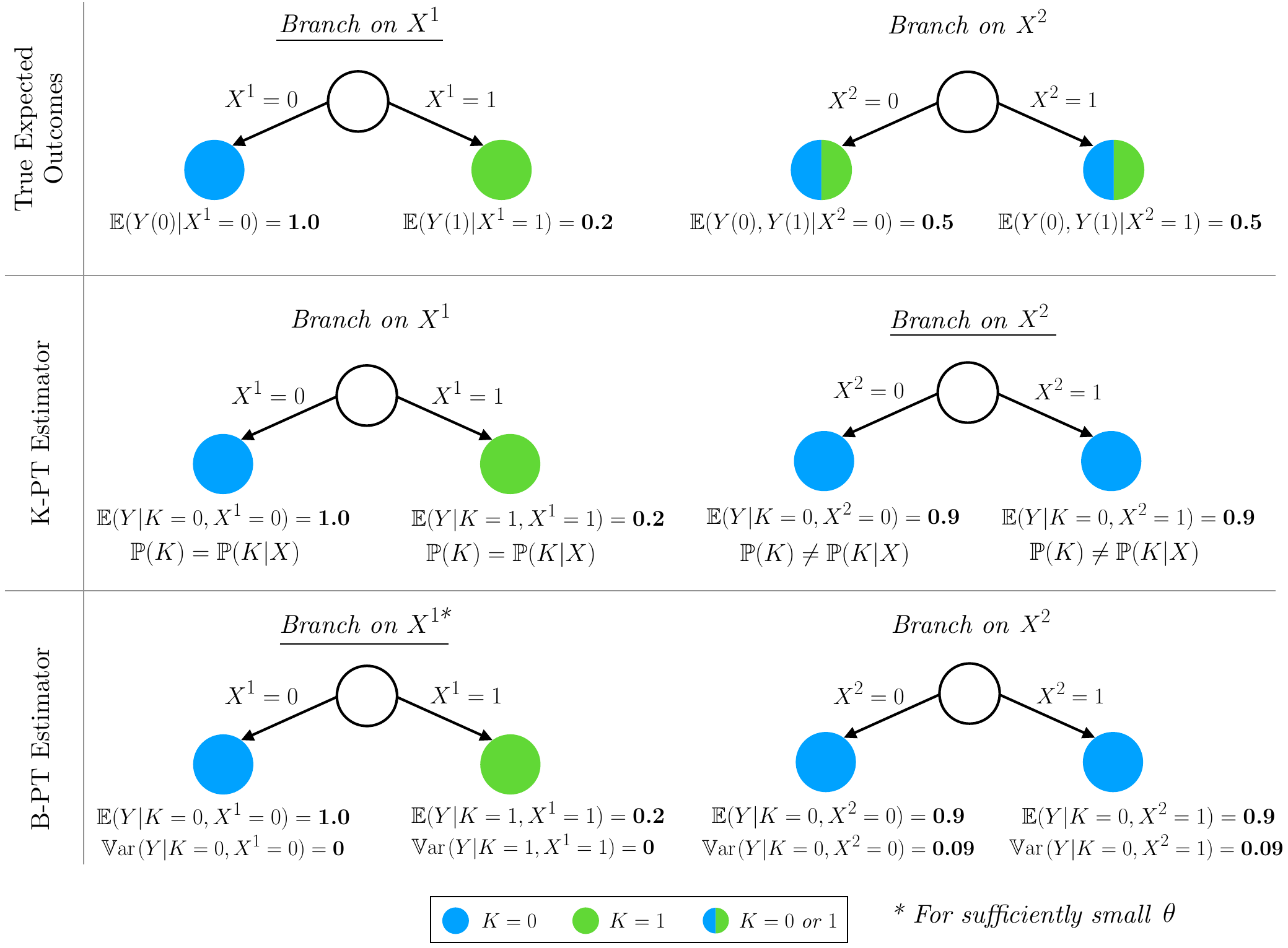}
	\centering
	\caption{Companion figure to Example~\ref{ex:example1}. Each row shows the estimates of performance of different prescriptive trees (branch on $X^1$--left-- or $X^2$--right--) under different estimators. In each case, only the best treatment assignments at each leaf (as measured by the corresponding estimator) are shown. In this example, the true optimal policy (top left) is to branch on $X^1$ and treat only those who are sick. Indeed, the expected outcome of the best tree branching on $X^1$ (top left) is $1.0 \cdot 0.5 + 0.2 \cdot 0.5 = 0.6$, which is larger than the expected outcome of the best $X^2$-branching tree (top right), which is $0.5 < 0.6$. However, the method from \cite{Kallus2017RecursiveData} (middle), which is discussed in Section \ref{section:kallus}, will prefer to branch on $X^2$ (noise covariate) and not treat any patient (middle right). Indeed, it estimates the outcomes of the best $X^1$- (resp.\ $X^2$-) branching tree as $0.6$ (resp.\ $0.9$). We note that Assumption~\ref{asu:sufficient_partition} is not satisfied at any of the leaves of the tree deemed as best by this estimator. In this case, the method of \cite{Bertsimas2019OptimalTrees}, see Section \ref{section:bertsimas}, is able to identify the truly optimal tree (bottom left); it correctly branches on~$X^1$ provided~$\theta$ is sufficiently small. Note that this method penalizes the variances corresponding to all treatments, but for simplicity the figure only depicts variances corresponding to the chosen treatment.}
	\label{fig:kallus_example}
\end{figure}

\revision{The optimal policies described above can be obtained by an omniscient decision-maker who knows the true counterfactuals $\mathbb{E}(Y(k))$. In practice, however, decision makers \revisio{can only obtain $\mathbb{E}(Y|K=k)$ from the observational data}. Naturally, incorrect estimates could potentially lead to poor decisions. In particular, we now argue that K-PT indeed produces suboptimal trees due to estimation \revisio{bias}.}
Consider the tree of depth one that maximizes the objective function of K-PT given by~\eqref{eq:kallus_estimator}. Suppose that we branch on covariate $X^j$ at the sole branching node of the tree, $j\in \{1,2\}$. Then, at the leaf where $X^j = x^j \in \{0,1\}$, K-PT estimates the counterfactual outcomes $\mathbb E(Y(k)|X^j=x^j)$ as $\mathbb{E}(Y|K=k, \; X^j=x^j)$. However, these estimates can be incorrect. For example,
$$
\mathbb{E}(Y(0)|X^2=0) = \sum_{x^1 \in \{0,1\}} \mathbb{E}(Y(0)|X^1=x^1,X^2=0)\mathbb{P}(X^1=x^1|X^2=0)=1 \times \frac{1}{2} + 0\times \frac{1}{2} = 0.5,
$$
whereas
\begin{equation*}
\begin{split}
\mathbb{E}(Y|K=0,X^2=0) = \sum_{x^1 \in \{0,1\}} \mathbb{E}(Y|K=0,X^1=x^1,X^2=0)\mathbb{P}(X^1=x^1|K=0,X^2=0) \\
= \newnj{\sum_{x^1 \in \{0,1\}} \mathbb{E}(Y|K=0,X^1=x^1) \frac{\mathbb{P}(K=0|X^1=x^1) \mathbb{P}(X^1=x^1)}{\mathbb{P}(K=0)}} 
=0.9. 
\end{split}
\end{equation*}
The difference between these two quantities is intuitive. Individuals that were not treated ($K=0$) in the data are more likely to be healthy (with a large associated outcome). The K-PT estimate $\mathbb{E}(Y|K=0,X^2=0)$ thus \emph{overestimates} $\mathbb{E}(Y(0)|X^2=0)=0.5$. Similarly, since the vast majority of individuals that were treated ($K=1$) in the data are sick (with low outcomes), the K-PT estimate $\mathbb{E}(Y|K=1,X^2=0)=0.26$ is an \emph{underestimator} for $\mathbb{E}(Y(1)|X^2=0)=0.5$. \newnj{Repeating this calculation for all combinations of $K$ and $X^2$ gives us $\mathbb{E}(Y|K=0, X^2=0)=\mathbb{E}(Y|K=0, X^2=1)=\mathbb{E}(Y|K=0)$ and similarly for $\mathbb{E}(Y|K=1, X^2=0)=\mathbb{E}(Y|K=1, X^2=1)=\mathbb{E}(Y|K=1).$ Hence, when splitting on $X^2$, the K-PT estimator will choose to not treat anyone ($K=0$) because the estimator's expected values are $0.9 > 0.26$, see the middle right subfigure of Figure~\ref{fig:kallus_example}.

We can conduct a similar analysis for the tree that branches on $X^1$. In this case, K-PT estimates $\mathbb{E}[Y|K=0, X^1=0] = 1$ precisely as stated in Table~\ref{table:expected}, since the historical policy assigns treatments solely based on feature $X^1$. Repeating the decisions for all combinations of $K$ and $X^1$ gives us a tree where K-PT will choose to treat ($K=1$) for those with $X^1=1$ and not treat otherwise, see the middle left subfigure of Figure~\ref{fig:kallus_example}. When choosing the optimal branching decision of the two options, K-PT will \textbf{incorrectly} choose to branch on $X^2$ because it estimates a higher expected outcome ($(0.9+0.9)/2 = 0.9$) than when branching on $X^1$ ($(1.0+0.2)/2 = 0.6$).

Note that in a depth 1 tree that only branches on $X^1$,} Assumption~\ref{asu:sufficient_partition} is satisfied and the estimated outcome of~$0.6$ provided by K-PT is correct. \revision{However, K-PT prefers the alternative tree which branches on the noise variable and does not satisfy Assumption~\ref{asu:sufficient_partition}, in this case resulting in an overly optimistic estimation of the outcomes.} We conclude that if not \emph{all feasible trees} in the problem satisfy Assumption~\ref{asu:sufficient_partition}, the method from~\cite{Kallus2017RecursiveData} may fail.

%

%
\label{ex:example1}
\end{example}


\subsection{B-PT}
\label{section:bertsimas}

\cite{Bertsimas2019OptimalTrees} also observe that K-PT may fail in certain conditions. They posit that failings can be attributed to poor estimates of the outcomes at the leaves. Thus, they augment the K-PT estimator with a regularization term, which penalizes trees that induce a high variance in the outcomes conditional on each treatment at the leaves, i.e., 
$$
\sum_{k \in \sets K} \sum_{n \in \sets L} \mathbb V{\rm{ar}} ( Y | X \in \sets X_n , K = k ).
$$
Specifically, they optimize the quantity
\begin{gather*}
\QbptI(\pi) = \theta \left(\QkptI(\pi)\right) - (1-\theta) \sum_{i \in \sets{I}} \left(Y_i - \widehat{Y}_{\sets{X}_{n(i)}}(K_i) \right)^2,
\end{gather*}
where $n(i) \in \sets L$ denotes the leaf datapoint $i$ belongs to, and $\theta\geq 0$ is a parameter to be tuned.

\subsubsection*{Analysis of B-PT.} Recall that \cite{Kallus2017RecursiveData}
estimates $\mathbb E(Y(k) | X \in \sets X_n)$ through $\mathbb E(Y | K=k, X \in \sets X_n)$ and shows that this estimator is unbiased under Assumption~\ref{asu:sufficient_partition}. By adding a regularization term that seeks to promote trees with low variance $\mathbb V{\rm{ar}}(Y| K=k, X \in \sets X_n)$, \cite{bertsimas2017personalized} improves the statistical properties of the K-PT estimator (lower variance). Therefore, if Assumption~\ref{asu:sufficient_partition} holds (so that $\mathbb{E}(Y|K=K_i, X\in \sets{X}_{n(i)})=\mathbb{E}(Y(K_i)| X\in \sets{X}_{n(i)})$), then B-PT can lead to improved performance over K-PT (for a suitable choice of $\theta$). However, B-PT does not directly address the issues discussed in Example~\ref{ex:example1}. Indeed, when Assumption~\ref{asu:sufficient_partition} does not hold, then $\mathbb{E}(Y|K=K_i, X\in \sets{X}_{n(i)})$ and $\mathbb{E}(Y(K_i)| X\in \sets{X}_{n(i)})$ can be drastically different from one another, and the policies chosen by B-PT may still be suboptimal.

In the specific case of Example~\ref{ex:example1}, \comment{it can be checked that if $\theta$ is sufficiently small, then} B-PT can in fact recover the best tree if $\theta$ is sufficiently small. Nonetheless, as we now show, a small modification of Example~\ref{ex:example1} results in a setting in which B-PT has \emph{additional} incentives to choose the worst tree, which branches on irrelevant features.  

\setcounter{example}{0}
\begin{example}[Continued]
We now revisit Example~\ref{ex:example1} and use the method from~\cite{Bertsimas2019OptimalTrees} to identify an optimal tree. Figure~\ref{fig:kallus_example} (bottom) shows the estimated outcomes and variances associated with each candidate tree. Since the tree that branches on the relevant feature~$X^1$ has smaller variance in this case, it will be chosen by B-PT provided that $\theta$ is sufficiently small. \hfill $\blacksquare$

\end{example}

\begin{example}
\label{ex:example2}
We now consider a variant of Example~\ref{ex:example1} where we add to all potential outcomes the quantity $(1-X^2)$, see Table~\ref{table:expected2}. \newnj{As before, $\mathbb P(K=0|X^1=0) = 0.9$ and $\mathbb P(K=0|X^1=1) = 0.1$, i.e., 90\% of healthy patients do not receive the treatment, and 90\% of sick patients do, also implying that $\mathbb{P}(K=0) =\mathbb{P}(K=1) = 0.5$. Since $X^2$ now affects the values of the outcomes, we have \revision{$\mathbb{E}(Y|K=0, X^2=1) \neq \mathbb{E}(Y|K=0, X^2=0)$} and similarly for $K=1$. However, this change does \textit{not}} affect the value $\mathbb E(Y(1)-Y(0)|X)$ and is thus \emph{irrelevant} in deciding which treatment is preferable. Therefore, under suitable identifiability conditions, there exists an optimal prescriptive tree of depth one\newnj{, which still splits on $X^1$}. Figure~\ref{fig:bertsimas_example2} (top) shows the outcomes, for a depth one tree, under the two possible branching decisions if the counterfactuals are known -- the best tree \newnj{indeed} branches on the only predictive feature $X^1$. However, when using estimates $\mathbb{E}[Y|K=k,X=x]$ for the counterfactuals (Figure~\ref{fig:bertsimas_example2}, bottom), the tree that branches on the irrelevant feature $X^2$ has both a higher estimated outcome and a smaller variance. \newnj{Therefore, B-PT will always choose to branch on $X^2$ and not treat anyone in the population, no matter the choice of $\theta$.} In this case B-PT has even more incentive to select a policy based on the irrelevant feature and both K-PT and B-PT will select a suboptimal tree. \hfill $\blacksquare$

\begin{table}[H]
\centering
 \begin{tabular}{c c | c c} 
 \hline
 $X^1$ & $X^2$ & $\mathbb{E}(Y(0))$ & $\mathbb{E}(Y(1))$ \\ [0.5ex] 
 \hline\hline
 0 & 0 & 2 & 1.8 \\ 
 \hline
 1 & 0 & 1 & 1.2 \\
 \hline
 0 & 1 & 1 & 0.8 \\
 \hline
 1 & 1 & 0 & 0.2 \\
 \hline
\end{tabular}
\caption{Companion table to Example~\ref{ex:example2}. Expectation of the potential outcomes in dependence of the covariate values. Similar to Example~\ref{ex:example1} (Continued), \newnj{we let the potential outcomes in Table~\ref{table:expected} have zero variance implying that potential outcomes conditioned on the covariate values are perfectly known}.}
\label{table:expected2}
\end{table}

\begin{figure}[h]
\includegraphics[width=\textwidth]{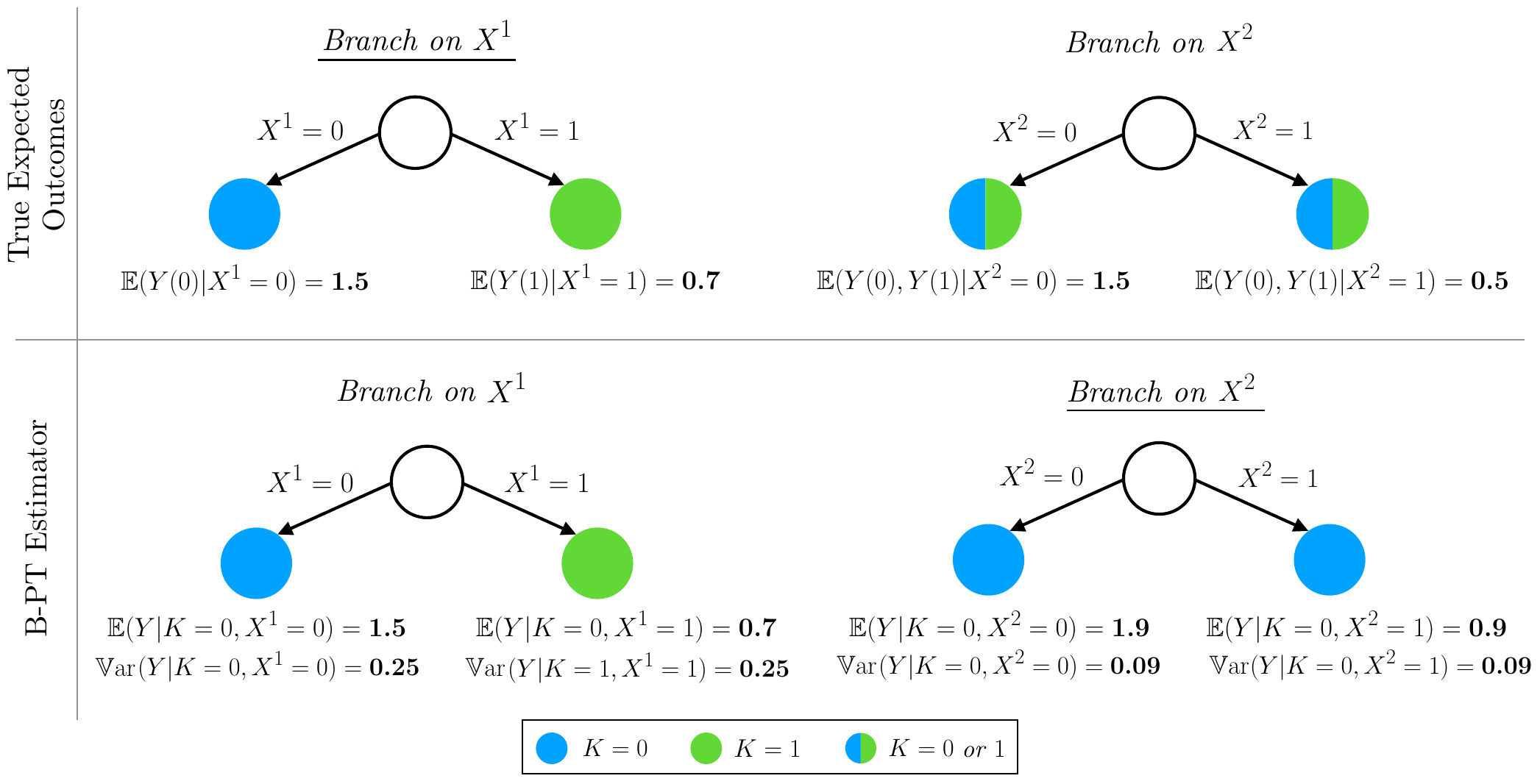}
\centering
\caption{Companion figure to Example~\ref{ex:example2}. The layout and interpretation of the figure exactly parallel that in Figure~\ref{fig:kallus_example}. The optimal tree (top left) has the objective $0.5(1.5) + 0.5(0.7) = 1.1$, which is higher than the $X^2$-branching tree (top right), $0.5(1.5) + 0.5(0.5) = 1.0 < 1.1$. However, B-PT will incorrectly branch on $X^2$ no matter the regularization strength $\theta$, since the $X^1$-branching tree (bottom left) has a lower estimated value and also higher variance.}
\label{fig:bertsimas_example2}
\end{figure}
\end{example}

\section{Proposed Formulations}
\label{section:proposed_formulations}

In this section, we present our proposed MIO formulations for learning optimal prescriptive trees from observational data. \newnj{We assume that the covariates are taken from a finite set of integers, i.e., $\sets{X} \subset \mathbb{Z}^F$, and let $\Theta(f)$ capture all levels of each feature $f \in \sets{F}$. \revision{To handle categorical features, one can one-hot encode the variables.} To handle continuous variables, one can pass in finite discretizations that are decided a priori. In Section~\ref{section:experimental_results}, we show that discretizing does not lead to a degradation in performance in our experiments; in fact, it may sometimes serve as a form of regularization and produce results that generalize  better out-of-sample than passing in continuous variables. We also note that discretization has a rich literature with many proposed solutions, see e.g., \cite{dougherty1995supervised,kotsiantis2006discretization, rucker2015researcher} for reviews of methods. }
Importantly, and in sharp contrast with the existing literature on prescriptive trees, we \emph{do not} make Assumption~\ref{asu:sufficient_partition}. Instead, we directly optimize the estimators~\eqref{eq:ipw_obj}, \eqref{eq:dm_obj}, or \eqref{eq:dr_obj} from the causal inference literature. This enables us to design interpretable prescriptive trees that are guaranteed to be optimal out-of-sample as the number of historical samples grows.

\begin{figure}[h]
\begin{center}
\centerline{
\includegraphics[width=0.45\textwidth]{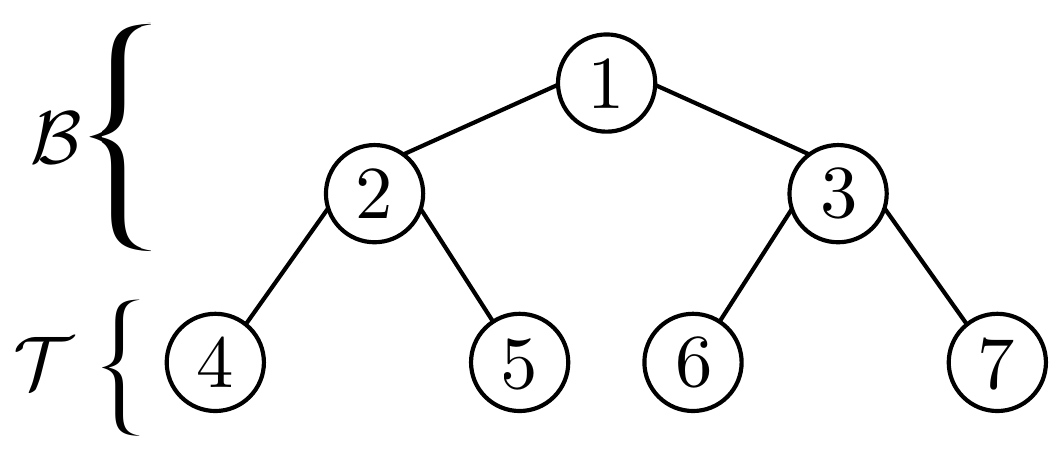}
\includegraphics[width=0.3\textwidth]{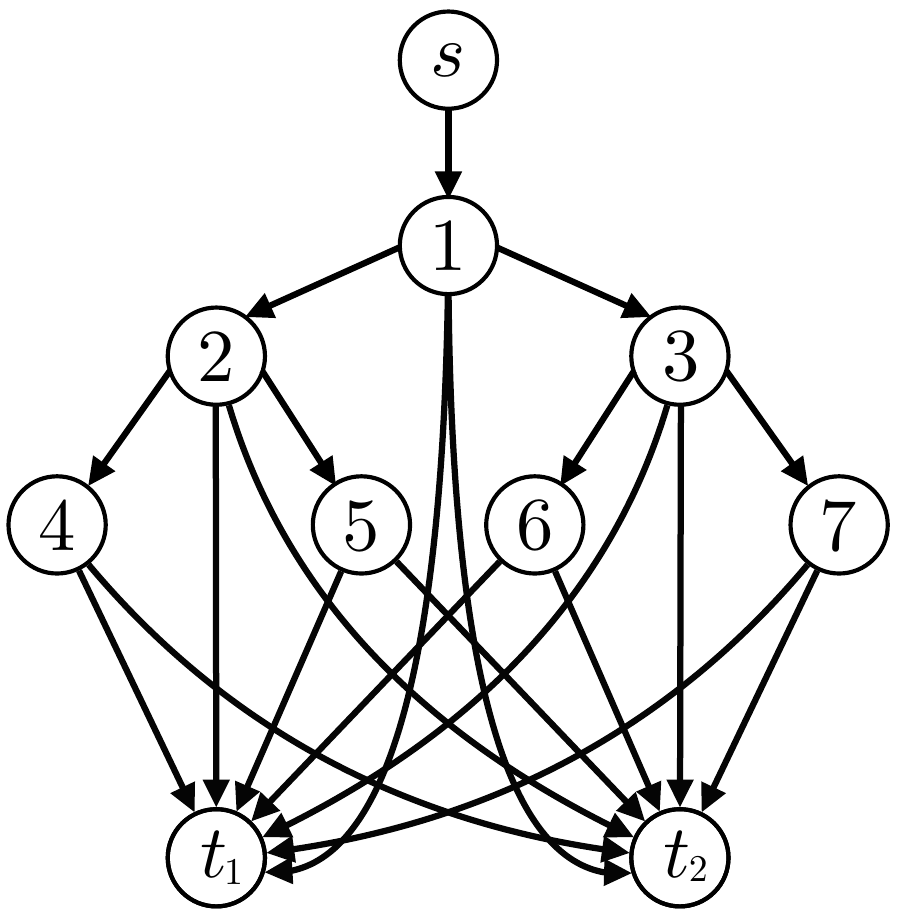}
}
\caption{A prescriptive tree with depth 2 (left) and its associated flow graph (right).}
\label{fig:flow_graph}
\end{center}
\end{figure}

As will become clear shortly, our proposed approaches give rise to \emph{weighted} classification trees. We thus adapt the MIO formulations of~\cite{aghaeistrong} for learning optimal classification trees from the predictive to the prescriptive setting. We extend our notation from the problem statement in Section \ref{section:problem_statement} to describe optimal prescriptive trees. The building block of our formulation is a perfect binary tree of depth~$d$ whose nodes are numbered $1$ through $(2^{d+1}-1)$ in the order in which they appear in a breadth-first-first search. \revision{Note that we can in general extend our method to learn non-binary trees similar to \cite{menickelly2016optimal}.} We let $\sets{B} := \{1, \dots, 2^{d}-1\}$ denote the set of branching nodes and $\sets{T} := \{2^d, \dots, 2^{d+1}-1\}$ collect all terminal nodes (see Figure \ref{fig:flow_graph}, left). From the perfect binary tree, we build a flow graph as follows. We connect a source node~$s$ to the root node, and connect all nodes other than $s$ to $|\sets{K}|$ sink nodes (denoted by $t_k, \forall k \in \sets{K}$) -- see Figure \ref{fig:flow_graph}, right. All links in the graph are directed from source to sink and have capacity 1.

\subsection{Formulation Based on Doubly Robust Estimation}
\label{section:robust_formulation}

Equipped with the flow graph associated with a prescriptive tree, we now formulate an MIO problem to optimize~\eqref{eq:dr_obj} over all trees of maximum depth $d$. For every branching node $n\in \sets B$, feature $f\in \sets F$, \newnj{and threshold $\theta \in \Theta(f)$,} we let the binary variable~$b_{nf\theta}$ indicate if feature~$f$ \newnj{with threshold $\theta$} is selected for branching at node~$n$. Accordingly, for every node~$n \in \sets{T} \cup \sets{B}$, we let the binary variable~$p_n$ indicate if node~$n$ is a treatment node, in which case a treatment must be assigned to all datapoints that land at that node and no further branching is allowed. For~$k\in \sets K$, we let~$w_{nk}\in \{0,1\}$ equal~1 if and only if (iff) treatment~$k$ is selected at node~$n$.

\newnj{In contrast to \cite{aghaeistrong} which decides the flow of each datapoint individually, we propose an \emph{aggregated} variant wherein we decide the flow of all datapoints with same covariates at once. Thus, we let $z^{a(n),n}_{x}$ indicate that datapoints with covariates $x \in \mathcal X$ will pass through arc $(a(n),n)$. This idea, based on the observation that datapoints with the same covariates follow the same path from root to sink, allows us to often sharply reduce the size of our formulations and speed up computation.}

The MIO formulation is as follows:
\begin{subequations}
\begin{align}
\text{maximize} \;\; & \newnj{\displaystyle \sum_{x \in \mathcal X} \sum_{k \in \sets K } \sum_{n \in \sets B \cup \sets T}
z^{n,t_{k}}_{x} \displaystyle \sum_{i: X_i = x} \left[\hat{\nu}_k(X_i) + \frac{\mathbbm{1}[k = K_i] (Y_i - \hat{\nu}_{K_i}(X_i))}{\mu(K_i, X_i)}\right]  }\label{eq:robust_obj}\\
\text{subject to}\;\;
& \displaystyle \sum_{f \in \sets F} \newnj{\sum_{\theta \in \Theta(f)} b_{nf\theta}} + p_n + \sum_{m \in \sets A(n)}p_{m} = 1   &\hspace{-5cm}  \forall n \in \sets B \label{eq:robust_branch_or_predict}\\
&  p_n+\sum_{m \in \sets A(n)}p_{m} =1   &  \forall n \in \sets T \label{eq:robust_all_terminal_leaf}\\
& \newnj{\displaystyle z^{a(n),n}_{x} =  z^{n,\ell(n)}_{x} + z^{n,r(n)}_{x} + \sum_{k \in \sets K}z^{n,t_{k}}_{x} }  &\hspace{-5cm}  \forall n \in \sets B,\; \newnj{x \in \mathcal X} \label{eq:robust_conservation_internal}\\
&  \newnj{\displaystyle z^{a(n),n}_{x} = \sum_{k \in \sets K}z^{n,t_{k}}_{x}}  &\hspace{-5cm}   \newnj{\forall x \in \mathcal X}, \; n \in \sets T \label{eq:robust_conservation_terminal}\\
& \newnj{\displaystyle z^{s,1}_{x} = 1} &\hspace{-5cm} \newnj{\forall x \in \mathcal X}\label{eq:robust_source}\\
&  \newnj{\displaystyle z^{n,\ell(n)}_{x}} \leq \sum_{f \in \sets F} \newnj{\sum_{\theta \in \Theta(f): x_{f}\leq \theta}}b_{nf\newnj{\theta}} &\hspace{-5cm} \forall n \in \sets B,\; \newnj{x \in \mathcal X} \label{eq:robust_branch_left}\\
&  \newnj{\displaystyle z^{n,r(n)}_{x}} \leq \sum_{f \in \sets F} \newnj{\sum_{\theta \in \Theta(f): x_{f} >\theta}}b_{nf\newnj{\theta}}  &\hspace{-5cm} \forall n \in \sets B, \; \newnj{x \in \mathcal X} \label{eq:robust_branch_right}\\
&  \newnj{\displaystyle z^{n,t_{k}}_x} \leq  w_{nk} &\hspace{-5cm} \forall \newnj{x \in \mathcal X},\; n \in \sets T,k \in \sets K \label{eq:robust_sink}\\
&  \displaystyle \sum_{k \in \sets K}w_{nk} = p_n  &\hspace{-5cm}  \forall n \in \sets B \cup \sets T \label{eq:robust_leaf_prediction}\\
&  \displaystyle w_{nk} \in \{0,1\}  &\hspace{-5cm}   \forall n \in \sets T, \; k \in \sets K \\
&  \displaystyle b_{nf\newnj{\theta}} \in \{0,1\}  &\hspace{-5cm}   \forall n \in \sets B, \; f \in \sets F, \; \newnj{\theta \in \Theta(f)} \\
&  \displaystyle p_{n} \in \{0,1\}  &\hspace{-5cm}   \forall n \in \sets B \cup \sets T \\
& \newnj{ \displaystyle z^{a(n),n}_{x}, z^{n,t_{k}}_{x}} \in \{0,1\}  &\hspace{-5cm}  \forall n \in \sets B \cup \sets T, \;\newnj{ x \in \sets X,} \; k \in \sets K. \label{eq:robust_dv_last}
\end{align}
\label{eq:robust}
\end{subequations}
Here, we have
where $\hat \mu$ and $\hat \nu_k$ are as defined in Sections~\ref{section:ipw_intro} and~\ref{section:dm}, respectively. $\sets A(n)$ denotes the set of all ancestors of node $n\in \sets B \cup \sets T$, $r(n)$ (resp.\ $\ell(n)$) is the right (resp.\ left) descendant of $n$. The objective \eqref{eq:robust_obj} maximizes the doubly robust objective~\eqref{eq:dr_obj}. Constraints~\eqref{eq:robust_branch_or_predict} ensure that we branch at all branching nodes where no treatment is assigned, at the node itself or at any of its ancestors. Similarly, constraints~\eqref{eq:robust_all_terminal_leaf} impose that a treatment must be assigned at all terminal nodes, unless a treatment has already been assigned to one of its ancestors. Constraints~\eqref{eq:robust_conservation_internal} and~\eqref{eq:robust_conservation_terminal} are flow conservation constraints, whereby any datapoint that flows into a node $n$ must either exit to its right or left descendant, or flow directly to one of the sink nodes. Constraints~\eqref{eq:robust_source} ensure that a flow of at most one can enter the source node for each datapoint. Constraints~\eqref{eq:robust_branch_left} and~\eqref{eq:robust_branch_right} guarantee that datapoints flow according to the branching decisions specified by variables $b$. Constraints~\eqref{eq:robust_sink} state that datapoints can only flow to the sink associated with their assigned treatment. Finally, constraints~\eqref{eq:robust_leaf_prediction} combined with integrality of the $w_{nk}$ variables ensure that if a treatment is assigned at a node, exactly one of the available treatments must be assigned. Note that we can relax integrality on the~$w$ and~$z$ variables and still reach an integral solution. \revision{We also note that in our formulation, $\sets{X}$ could comprise only the features we observe in the data, rather than the entire feature space.}

\newnj{\subsection{Formulation Based on Inverse Propensity Weighting Estimator}
\label{section:ipw_formulation}

To optimize the inverse propensity weighting objective~\eqref{eq:ipw_obj}, we can replace~\eqref{eq:robust_obj} to get 

\begin{equation}\label{eq:ipw_mio}
\begin{split}
\text{maximize} \;\;\;\;\; & \displaystyle \sum_{x \in \mathcal X} \sum_{k \in \sets K } \sum_{n \in \sets B \cup \sets T}
z^{n,t_{k}}_{x} 
\displaystyle \sum_{i: X_i = x} \frac{\mathbbm{1}[K_i = k] Y_i}{\mu(K_i, X_i)} \\
\text{subject to} \;\;\;\;\; & \eqref{eq:robust_branch_or_predict}\text{-}\eqref{eq:robust_dv_last}.
\end{split}
\end{equation}

\subsection{Formulation Based on Direct Method Estimator}
\label{section:dm_formulation}
Formulation~\eqref{eq:robust} can similarly be adapted to optimize the direct method objective~\eqref{eq:dm_obj} by simply dropping the second term in the objective~\eqref{eq:robust_obj}, i.e.,
\begin{equation}\label{eq:dm_mio}
\begin{split}
\text{maximize}  \;\;\;\;\; & \displaystyle \sum_{x \in \mathcal X} \sum_{k \in \sets K } \sum_{n \in \sets B \cup \sets T} \displaystyle
z^{n,t_{k}}_{x}  \displaystyle \sum_{i: X_i = x} \hat{\nu}_k(X_i) \\
\text{subject to} \;\;\;\;\; & \eqref{eq:robust_branch_or_predict}\text{-}\eqref{eq:robust_dv_last}.
\end{split}
\end{equation}}

\subsection{{Analysis of IPW, DR, and DM}}
\label{section:analysis}

It can be shown that, \newsa{for the data presented in Examples~\ref{ex:example1} and~\ref{ex:example2}, if $\mu$ and/or $\nu$ are known, a depth one tree trained according to formulations~\eqref{eq:robust},~\eqref{eq:ipw_mio}, and~\eqref{eq:dm_mio} will be optimal out-of-sample,} provided the number of samples in the data is sufficiently large. In this section, we go one step further and demonstrate that, under \newsa{suitable} conditions \newpv{on the historical data generation process and the estimators}, \newsa{the tree returned by our optimal MIO formulations~\eqref{eq:robust},~\eqref{eq:ipw_mio}, and~\eqref{eq:dm_mio} will converge almost surely to an optimal tree.}

We introduce some notation to help formalize our claim. 
For any tree based policy $\pi \in \Pi_d$, define 
$$\Qipw (\pi) := \mathbb E\left[ \frac{ \mathbb I (K= \pi(X)) Y }{ \mu (K,X) } \right] \quad \text{and} \quad \QipwIN (\pi) := \frac{1}{I} \sum_{i=1}^{I} \frac{ \mathbb I (K_i= \pi(X_i)) Y_i }{ \muhatN (K_i,X_i) },$$
where $\muhatN (K,X)$ is an estimator of $\mu (K,X)$ trained over a\newsa{n i.i.d.} sample of cardinality $N$ independent from~$\mathcal D$.
We denote by~$v^\star$ and~$\sets S^\star$, the optimal value and the set of optimal solutions of~$\max_{\pi \in \Pi_d} Q(\pi)$, respectively. 
Similarly, we denote by~$\vhatINipw$ and~$\shatINipw$, the optimal value and the set of optimal solutions of~$\max_{\pi \in \Pi_d} \QipwIN(\pi)$, respectively. Finally, we use the abbreviation ``w.p.1" for ``with probability one". 
\begin{proposition}
\label{prop:exact_IPW} 
Suppose that: conditional exchangeability holds, i.e., $Y(k) \independent K | X$, for all values of~$k \in \sets K$; the potential outcomes $Y(k)$ are bounded, for all values of~$k \in \sets K$; and, both $\mu(k,x) >0$ and $\muhatN(k,x)>0$ for all $x \in \sets X$, $k\in \sets K$, and $N$ sufficiently large. If $\muhatN\newsa{(k,x)}$ converges almost surely to~$\mu\newsa{(k,x)}$ \newsa{for all $x \in \sets X,~k \in \sets K$,} then $\vhatINipw \rightarrow v^\star$ w.p.1 and $\shatINipw \subseteq \sets S^\star$ w.p.1 for $I$ and $N$ large enough.
\end{proposition}

Similar to the IPW case, we now show that under \newsa{suitable} conditions, the MIO formulation~\eqref{eq:robust} for the DR estimator returns an optimal out-of-sample tree. We introduce additional notation to help formalize our claim.
For any tree based policy $\pi \in \Pi_d$, define
$$\Qdr (\pi) \; := \; \mathbb E\left[ \hat{\nu}_{\pi(X)}(X) + (Y-\hat{\nu}_{\pi(X)}(X))\frac{ \mathbb I (K= \pi(X)) }{ \hat{\mu} (K,X) } \right]$$
$$\text{and} \quad \QdrIN (\pi)\; := \; \frac{1}{I} \sum_{i=1}^{I} \left(\nuhatN_{\pi(X_i)}(X_i) + (Y-\nuhatN_{\pi(X_i)}(X_i))\frac{ \mathbb I (K_i= \pi(X_i)) }{ \muhatN(K_i,X_i) }\right),$$
where $\nuhatN_K (\cdot)$ is an estimator of~$\nu_K(\cdot)$, trained over a\newsa{n i.i.d.} sample of cardinality~$N$ independent from~$\mathcal D$. Let $\hat{\mu}(K,X)$ (resp. $\hat{\nu}_K(X)$) be the limit of $\muhatN (K,X)$ (resp. $\nuhatN_K (X)$) as $N$ goes to infinity.
We denote by~$\vhatINdr$ and~$\shatINdr$, the optimal value and the set of optimal solutions of~$\max_{\pi \in \Pi_d} \QdrIN(\pi)$, respectively.

\begin{proposition}
\label{prop:exact_DR}
Suppose that: conditional exchangeability holds, i.e., $Y(k) \independent K | X$, for all values of~$k \in \sets K$; the potential outcomes $Y(k)$ are bounded, for all values of~$k \in \sets K$;  both $\mu(k,x) >0$ and $\muhatN(k,x)>0$ for all $x \in \sets X$, $k\in \sets K$ and $N$ sufficiently large; and both $\hat \mu(k,x)$ and $\hat \nu_k (x)$ exist and are bounded for all $x \in \sets X$, $k\in \sets K$. If either $\muhatN\newsa{(k,x)}$ converges almost surely to $\mu\newsa{(k,x)}$ or $\nuhatN_k\newsa{(x)}$ converges almost surely to $\nu_k\newsa{(x)}$ \newsa{for all $x \in \sets X,~k \in \sets K$}, then $\vhatINdr \rightarrow v^\star$ w.p.1 and $\shatINdr \subseteq \sets S^\star$ w.p.1 for $I$ and $N$ large enough.
\end{proposition}

At the end, we show that under \newsa{suitable} conditions, the MIO \newsa{formulation~\eqref{eq:dm_mio}} returns an optimal out-of-sample tree. We introduce additional notation to help formalize our claim.
For any tree based policy $\pi \in \Pi_d$, define
$$\Qdm (\pi) \; := \; \mathbb E\left[ \hat{\nu}_{\pi(X)}(X) \right] \quad \text{and} \quad  \QdmIN (\pi)\; := \; \frac{1}{I} \sum_{i=1}^{I} \left(\nuhatN_{\pi(X_i)}(X_i)\right).$$
We denote by~$\vhatINdm$ and~$\shatINdm$ the optimal value and the set of optimal solutions of~$\max_{\pi \in \Pi_d} \QdmIN(\pi)$, respectively.

\begin{proposition}
\label{prop:exact_DM}
Suppose that: conditional exchangeability holds, i.e., $Y(k) \independent K | X$, for all values of~$k \in \sets K$; and $\hat \nu_k (x)$ exists, for all $k \in \sets K$. If $\nuhatN_k(x)$ converges almost surely to $\nu_k(x)$ \newsa{for all $x \in \sets X$, $k \in \sets K$}, then $\vhatINdm \rightarrow v^\star$ w.p.1 and $\shatINdm \subseteq \sets S^\star$ w.p.1 for $I$ and $N$ large enough.
\end{proposition}

\subsection{Extensions}\label{sec:extensions} \newnj{Our formulations in Sections \ref{section:robust_formulation}, \ref{section:ipw_formulation}, and \ref{section:dm_formulation} have strong modeling power. For example,} they can be used to learn trees that interpretable and/or fair, to design randomized policies, and to impose budget constraints relevant for under-resourced settings. We discuss these in the present section.

\subsubsection*{Intepretability.} In settings where learning interpretable policies is desired, see e.g., \cite{rudin2019stop} and~\cite{azizi2018designing}, one may limit the number of branching nodes in the tree to $M$ by augmenting formulations~\eqref{eq:robust}\newnj{, \eqref{eq:ipw_mio}, and \eqref{eq:dm_mio}} with the constraint $$\sum_{n\in \sets B\cup \sets T}p_n \; \leq \; M.$$

\subsubsection*{Budget Constraints.} In low resource settings such as organ allocation \citep{bertsimas2013fairness, zenios2000dynamic}, housing allocation \citep{azizi2018designing}, and security \citep{xu2018strategic}, many treatments/interventions often have limited supply. Tree based policies that satisfy such capacity constraints can be learned by augmenting formulations~\eqref{eq:robust}\newnj{, \eqref{eq:ipw_mio}, and \eqref{eq:dm_mio}} with constraints

$$ \displaystyle \sum_{n \in \sets B \cup \sets T} \newnj{\sum_{x \in \sets X} \sum_{i \in \sets{I}: X_i = x} z^{n,t_{k}}_x } \; \leq \; |\sets{I}|C_k \quad\forall k \in \sets K, $$
where $C_k$ denotes the percentage of instances that can be assigned treatment $k$.

\subsubsection*{Fairness in Treatment Assignment.} Our formulations can be used to learn trees where treatments are assigned fairly across groups. Concretely, we will say that a policy satisfies treatment assignment parity if the probability of assigning a particular treatment is equal across protected groups. \newnj{This constraint is similar to notions of fairness in the classification setting, for which there is a rich literature -- see, e.g., \cite{dwork2012fairness, zemel2013learning, bolukbasi2016man, hardt2016equality, zafar2017fairness, olfat2018spectral, jo2022fairness}.} We let $\sets P$ collect \newnj{all protected covariates (e.g., LGBTQ status or categories of race)}. Accordingly, we let \newnj{$P_x \in \sets P$} represent the value of the protected feature(s) of \newnj{covariate~$x$}. It is worth noting that the features in~$\sets P$ are usually not included in covariates~$X$ to make sure that the policy does not make decisions based on protected features. To ensure that the learned policy satisfies treatment assignment parity up to a bias $\delta$, we may augment~\eqref{eq:robust}\newnj{, \eqref{eq:ipw_mio}, and \eqref{eq:dm_mio}} with the constraint
$$\left| \frac{\displaystyle \sum_{n \in \sets B \cup \sets T} \displaystyle \newnj{\sum_{x \in \sets X : P_x = p} \sum_{i \in \sets{I}: X_i = x} z^{n, t_k}_x}}{|\{i \in \sets I : P_i = p\}|} - \frac{\displaystyle \sum_{n \in \sets B \cup \sets T} \newnj{\displaystyle \sum_{x \in \sets X : P_x = p'} \sum_{i \in \sets{I}: X_i = x}  z^{n, t_k}_x}}{|\{i \in \sets I : P_i = p'\}|} \right| \; \leq \; \delta \qquad \forall p, p' \in \sets P: p \neq p', \; k \in \sets K.$$

In a similar fashion, we will say that a policy satisfies \textit{conditional} treatment assignment parity if the probability of assigning a particular treatment is equal across protected groups, conditional on some legitimate feature(s) that affect the outcome. For instance, in prescribing housing resources to people experiencing homelessness, one may require that all individuals with the same vulnerability have the same likelihood of receiving a certain treatment.  We let $\sets A$ be the set of features indicative of risk, and let $A_i \in \sets A$ be the value of the risk feature of datapoint $i$. Conditional treatment assignment parity is satisfied up to a bias $\delta$ and for all features $a \in \sets A$ by adding the following constraint to \eqref{eq:robust}\newnj{, \eqref{eq:ipw_mio}, and \eqref{eq:dm_mio}},
\begin{equation*}
\begin{split}
\left| \frac{\displaystyle \sum_{n \in \sets B \cup \sets T} \newnj{\displaystyle \sum_{x \in \sets X : P_i = p, A_x = a} \sum_{i \in \sets{I}: X_i = x}  z^{n, t_k}_x}}{|\{i \in \sets I : P_i = p \cap A_i = a\}|} - \frac{\displaystyle \sum_{n \in \sets B \cup \sets T} \newnj{\displaystyle \sum_{x \in \sets X : P_x = p', A_x = a} \sum_{i \in \sets{I}: X_i = x}  z^{n, t_k}_x}}{|\{i \in \sets I : P_i = p' \cap A_i = a \}|} \right| \; \leq \; \delta  \\ \forall p, p' \in \sets P: p \neq p', \; k \in \sets K, \; a \in \sets A.
\end{split}
\end{equation*}

\subsubsection*{Fairness in Treatment Outcomes.} Our formulations can also be used to ensure fairness in expected outcomes across groups. In general, this requires the average expected outcomes of a protected group $p \in \sets P$ to be above some threshold~$\gamma_p$ $\in \mathbb R$:

$$\displaystyle \sum_{n \in \sets B \cup \sets T} \displaystyle \newnj{\sum_{x \in \sets X : P_x = p}} \displaystyle \sum_{k \in \sets{K}} 
\newnj{\displaystyle \sum_{i: X_i = x} z^{n,t_{k}}_{x} \left[ \hat{\nu}_k(X_i) + \frac{\mathbbm{1}[k = K_i] (Y_i - \hat{\nu}_{K_i}(X_i))}{\mu(K_i, X_i)} \right]} \; \geq \; \gamma_p \qquad \forall p \in \sets P.$$

%
For example, \cite{bertsimas2013fairness} set
$\gamma_p := \sum_{i \in \sets I: P_i = p} Y_i$, i.e., the expected outcomes of all protected groups under the learned policy should be greater than or equal to what was observed in the data. Alternatively, the constraint can be used to impose max-min fairness to protect the outcomes of the groups that are worst off, see \cite{rawls1974some}. In this case, one sets $\gamma_p := \gamma$ for all $p \in \sets P$, where $\gamma$ is the largest value for which the MIO problem remains feasible.

\subsubsection*{Randomized Treatment Assignment Policies.} The addition of fairness and budget constraints may make problems~\eqref{eq:robust}, \newnj{\eqref{eq:ipw_mio}, and \eqref{eq:dm_mio}} infeasible. In such cases, it may be desirable to design \emph{randomized} policies where datapoints that land at the same leaf are assigned each treatment with a certain probability (rather than all getting the same treatment). This can be achieved by relaxing integrality on the variables~$w$ and~$z$ in formulations~\eqref{eq:robust}\newnj{, \eqref{eq:ipw_mio}, and \eqref{eq:dm_mio}}. The variable $w_{nk}$ can then be interpreted as the probability of assigning treatment~$k$ to datapoints that fall on node~$n$.

\section{Experiments}
\label{section:experiments}

We now evaluate the empirical performance of our proposed formulations in Section~\ref{section:proposed_formulations} in two problem settings: a synthetic setting from the literature and a real setting based on a warfarin dosing dataset. In both cases, we take the viewpoint that data comes from an observational study, which can be viewed as a conditionally randomized experiment. We benchmark against the approaches of~\cite{Kallus2017RecursiveData} and \cite{Bertsimas2019OptimalTrees} for learning prescriptive trees, see Section~\ref{section:prev_mio}. As before, we refer to these as K-PT and B-PT, respectively. \revisio{We also compare our method to four other, non-MIO based, approaches that can similarly learn or be adapted to learn prescriptive policies: causal forests and causal trees \citep{Athey2016RecursiveEffects}, policy tree \citep{zhou2023offline}, and regress \& compare (see Section~\ref{section:dm}). We describe these methods further in Section~\ref{sec:other_methods}.} Since \cite{Bertsimas2019OptimalTrees} do not propose an MIO formulation, we adapt the method from \cite{Kallus2017RecursiveData} to implement B-PT. For completeness, we provide the MIO formulations that we have implemented for these two approaches in Appendices~\ref{section:kallus_formulation} and~\ref{section:bertsimas_formulation}, respectively. We evaluate the performance of all approaches as the probability of correct treatment assignment is varied in the historical data.

\subsection{Dataset Description}
\label{sec:dataset_description}

\subsubsection*{Synthetic Data.} For our experiments on synthetic data, we adapt the data generation process from \cite{Athey2016RecursiveEffects}. In this problem, there are two treatment possibilities indexed in the set $\sets K = \{0, 1\}$. The covariate vector has two independent and identically distributed features, $X=(X^1, X^2)$, where each $X^j \sim \mathcal{N}(0, 1)$. The potential outcome of datapoint~$i$ with covariates~$X_i$ under treatment~$k$ is
\begin{equation}\label{eq:synthetic}
Y_i(k) = \phi(X_i) + \frac{1}{2}(2k-1) \cdot \kappa(X_i) + \epsilon_i,
\end{equation}
where $\phi(x) := \frac{1}{2}x^1 + x^2$ models the mean effect, $\kappa(x):=\frac{1}{2}x^1$ models the treatment effect, and $\epsilon_i \sim \sets N(0,0.1)$ is noise added to the outcome that is independent of the covariates.

To study different settings of observational experiments, we vary, in the data, the probability~$p$ of assigning the treatment that is best \emph{in expectation} for each unit\comment{; we denote this probability by~$p$}. We let $p \in \{ 0.1, 0.25, 0.5 , 0.75, 0.9\}$ -- since there are 2 treatments, 0.5 corresponds to the (marginally) randomized setting, while the other settings correspond to conditionally randomized experiments. For each~$p$, we randomly generate 5 training and test sets, each with 500 and 10,000 datapoints, respectively. \newnj{We use the training set to estimate $\mu$ and $\nu$, and to train formulations~\eqref{eq:robust},~\eqref{eq:ipw_mio}, and~\eqref{eq:dm_mio}; we use the test set to evaluate the learned policy. Since this is a synthetic dataset, we have access to counterfactuals in the test set, which we use for evaluation purposes.}

\subsubsection*{Warfarin Dosing.} For our study on real data, we employ a dataset for personalized warfarin dosing for which counterfactuals are available. Warfarin is the most widely used oral anticoagulant agent, but despite its prevalence, determining one's optimal warfarin dosage is difficult because it can vary widely depending on demographic variables, clinical factors, and genetics \citep{international2009estimation}. The publicly available dataset that we \comment{employ}use was collected by the International Warfarin Pharmacogenetics Consortium and published at the Pharmacogenetics and Pharmacogenomics Knowledge Base, see \cite{international2009estimation}. The advantage of using this dataset is that we can model a patient's true outcomes when given varying doses of warfarin, which allows us to evaluate the performance of arbitrary counterfactual policies. \cite{international2009estimation} published a learned affine function $f(x) = \beta x + c$ that determines the optimal warfarin dose based on a patient's age, weight, race, VKORC1 genotype, CYP2c9 genotype, and whether or not the patient is currently taking amiodarone or an enzyme reducer, see equation~\eqref{warfarin_function} in Section~\ref{sec:appendix_warfarin}. We calculate a patient $i$'s optimal dosage using $f(X_i) + \epsilon$, where $\epsilon \sim \mathcal{N}(0, 0.02).$ This dosage is then discretized into three groups (i.e., $|\mathcal K|=3$) using the same convention as \cite{international2009estimation}: $K_{i}^{\text{opt}} = 0$ ($\leq$ 3 mg/day), 1 (between 3 and 7 mg/day), and 2 ($\geq$ 7 mg/day). The observed outcome is $Y_{i}(K_i) = 1$, if $K_i = K_{i}^{\text{opt}}$; and $=0$ otherwise.

To study different settings of observational experiments, we consider three different treatment assignment mechanisms, as follows. For the marginally randomized setting, we assign each treatment with probability $1/3$, and repeat to produce 5 datasets. To simulate data that is based on a more informed policy, we assign treatments based on modified versions of $f$ obtained by perturbing its coefficients so that patients closer to the boundary of each treatment bucket are at higher risk of receiving an incorrect treatment. Specifically, for each coefficient~$a$ of $f$, we generate $a'\sim U(a-a\cdot r, a+a \cdot r)$, where $r$ denotes the range of sampling. We assign patient $i$ the treatment corresponding to this modified function, discretized into three groups as discussed previously. In testing, we found that $r \in [0.05, 0.12]$ was the ideal range to simulate a reasonable treatment policy where the probability of correct treatment assignment is between 0.6 and 0.9. We fix 2 values of~$r$, 0.06 and 0.11, and for each value, we generate 5 sets of randomly sampled coefficients (from hereon, we refer to this variability as a ``realization''). At this point we have~3 experiment designs --each with~5 realizations -- making up~15 datasets. Each dataset is then split randomly~5 times into training and test sets of~3,000 and~1,386 instances, respectively, yielding a total of~75 train-test pairs. \newnj{Similar to the synthetic data, only the training set is used to estimate $\mu, \nu$, and to train formulations~\eqref{eq:robust},~\eqref{eq:ipw_mio}, and~\eqref{eq:dm_mio}.}

\subsection{Experimental Setup}\label{sec:experimental_setup}
Since the proposed methods rely on estimating the propensity scores and/or counterfactuals, we discuss the estimation models we use for both datasets. We also discuss the preprocessing steps required. \revisio{All approaches are allotted a solve time of 4 hours and utilize 6 Intel Xeon E5-2640 v4, 2.40GHz CPUs, each having 4GB of memory. The MIO-based approaches use the Gurobi\footnote{See \url{https://www.gurobi.com/products/gurobi-optimizer/}} solver (version 10.0.0).}

\subsubsection*{Synthetic Data.} The synthetic data has real covariates but our methods only allow for binary features to split on. We thus discretize each covariate feature into ten buckets corresponding to deciles from a normal distribution. Since the outcome distribution is known by construction (it is linear), we learn the potential outcomes using both linear regression (LR) and lasso regression with $\alpha = 0.08$ (Lasso), knowing that the latter will be a slightly less accurate predictor. We also use true propensity scores--they are known by construction--as well as two methods to estimate propensity scores: logistic regression (Log) and decision trees (DT). Therefore, there will be six doubly robust methods, each corresponding to a pair of models for IPW and DM.

\subsubsection*{Warfarin Dosing.} Most of the features are already binary with the exception of age, height, and weight, so these features were split into 5 buckets, where each bucket contains approximately the same number of datapoints. A variety of other preprocessing and data imputation steps were done based on recommendations from \cite{international2009estimation}. A patient's possible outcomes are binary \revision{and thus we use the predicted class probabilities from $\hat{\nu}$ to train our trees}. \newnj{Recall that in order to predict counterfactual outcomes for a given treatment option, we can only train a predictive model on the datapoints that were given said treatment in the historical policy; therefore, we evaluate a model's performance on how well it generalizes on the entire population -- particularly, 
we measure mean square loss and AUC. We find that, depending on the data generation process, different models \revision{of $\hat{\nu}$} yield the best performance in the population: for the marginally randomized setting, random forests (RF) with balanced class weights were best; for the non-randomized settings, a combination of weighted RFs and logistic regression (LR) were best to account for class imbalance. The propensity scores are learned by fitting an ML model to predict historical treatment assignment $K$ from covariates $X$ -- we use decision trees (DT) because it yields the best performance in the population (lowest mean square loss and highest AUC)}.

\subsubsection*{Policy Evaluation.} We evaluate a learned policy~$\pi$ in terms of its out-of-sample probability of assigning the optimal treatment to a patient (OOSP). In the next section, we compare all methods using this metric.

\newnj{
\subsection{Other Benchmark Methods from the Literature}\label{sec:other_methods}
\subsubsection*{Regress and Compare (R\&C).} We employ the regress and compare approach as described in Section~\ref{section:dm}, which serves as a highly personalized (but uninterpretable) policy. For the experiments on synthetic data, we use linear regression to match the potential outcome functions~\eqref{eq:synthetic}. For the experiments on warfarin dosing, we use three models: \textit{1)} logistic regression (LR) with balanced class weights, \textit{2)} random forests (RF) with balanced class weights, and \textit{3)} the model (either LR and RF) with custom class weights resulting in the best performance in the population for each dataset, as described in Section~\ref{sec:experimental_setup}.

\subsubsection*{Causal Forests and Causal Trees (CF \& CT).} \cite{Athey2016RecursiveEffects} propose a method that adapts the CART and random forest algorithm to learn heterogeneous treatment effects in a population. Each causal tree recursively partitions in a way that maximizes differences in treatment effects across splits (typically, difference between outcomes of the treatment and control groups). A causal forest grows hundreds to  thousands of causal trees and aggregates results across the trees. However, CF and CT are concerned with estimation (rather than prescription) and they only consider binary treatment options. To produce a fairer comparison with our method, we propose the following approach, which was adapted from~\cite{Kallus2017RecursiveData}. We first designate a baseline treatment option $k_0$ (e.g., $k_0=0$). For all other treatment options $k \in \sets{K} \setminus \{ k_0 \}$, we use CF or CT to estimate $\delta^k(x) := \mathbb{E}[Y|K=k, X=x] -\mathbb{E}[Y|K=k_0, X=x]$ (which we denote by $\hat{\delta}^k(x))$. We also let $\hat{\delta}^{k_0}(x)$ be 0 for all $x$. Finally, we prescribe treatments via $\argmin_{k \in \sets{K}} \hat{\delta}^k(x)$ to all individuals $x \in \sets{X}$. We implement both CF and CT using the grf package in R\footnote{\url{https://github.com/grf-labs/grf}}. 

\subsubsection*{Policy Tree (PT).} \cite{zhou2023offline} propose to learn optimal prescriptive trees using a similar doubly robust objective. One critical difference from our method is that PT learn their trees recursively rather than using MIO. \revision{Therefore, while PT can learn trees with continuous data and in less time, it cannot handle additional constraints} (see Section~\ref{sec:extensions}). As such, to further showcase the flexibility of our method, we will run experiments incorporating budget and fairness constraints in Section~\ref{sec:additional_experiments}. We implement PT via the R package policytree\footnote{\url{https://github.com/grf-labs/policytree}} without discretizing the features.
}

\subsection{Experimental Results}
\newnj{We now analyze the results for our method in relation to the aforementioned works. We first discuss our experiments on the synthetic data, and then move to a discussion on the warfarin experiments.}


\label{section:experimental_results}
\begin{figure}[h]
\includegraphics[width=0.85\textwidth]{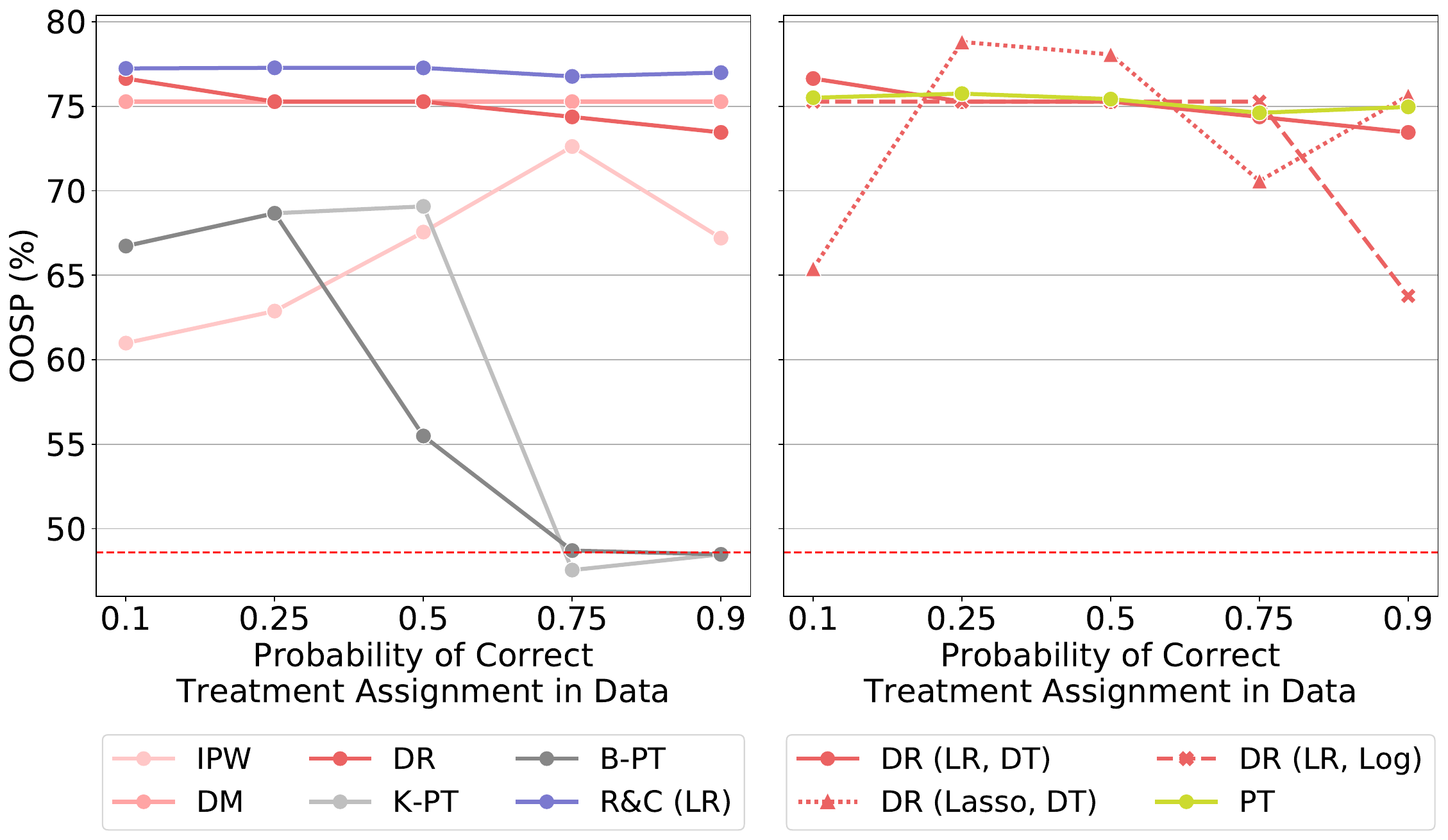}
\centering
\caption{\newnj{Results on the synthetic dataset. The left graph compares the out-of-sample probability of correct treatment assignment of the best models for IPW, DM, and DR with a linear regressor regress \& compare (R\&C) approach, as well as methods described in \cite{Kallus2017RecursiveData} (K-PT) and \cite{Bertsimas2019OptimalTrees} (B-PT). The right graph shows the method described in \cite{zhou2023offline} (PT), as well as our DR methods where at least one of the combined methods is correct, i.e., linear regression (LR) for estimating $\nu$ and decision trees (DT) for estimating $\mu$. Lasso regression (Lasso) and logistic regression (Log) are both suboptimal models. Both figures are averages from trees of depth $d = 1$ with 2 leaf nodes.}}
\label{fig:athey_v1}
\end{figure}

\begin{figure}[h]
\includegraphics[width=0.6\textwidth]{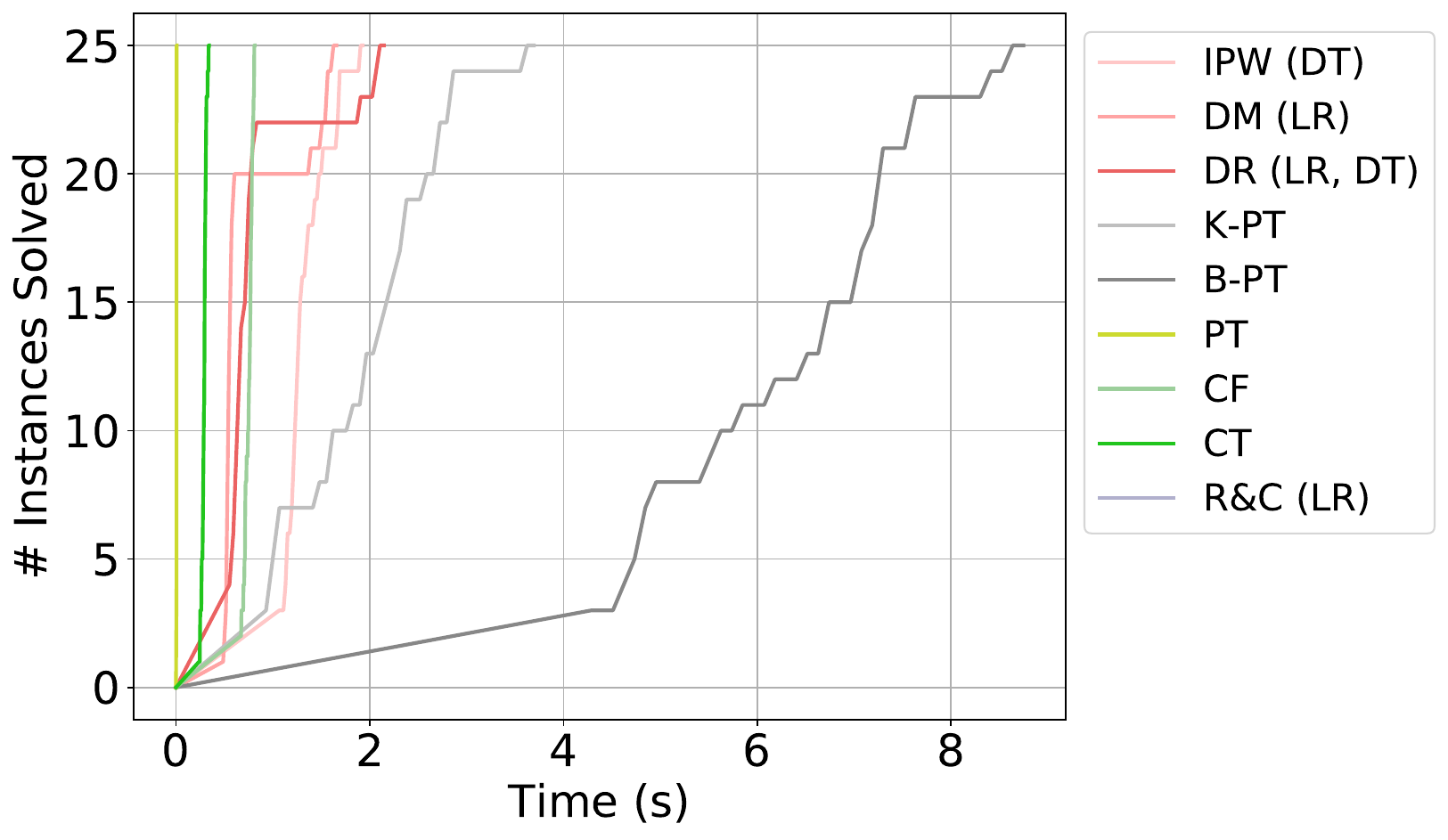}
\centering
\caption{\newnj{Comparison of computational performance for the synthetic data. All methods solved to optimality in less than 9 seconds. We exclude DR (Lasso, DT) and DR (LR, Log) for brevity because the computational times are similar to DR (LR, DT), the best performing model.}
\label{fig:synthetic_comp_times}}
\end{figure}

\newnj{
\subsubsection{Synthetic Data}\label{sec:exp_results_synthetic}

\paragraph{Optimization Performance.}
OOSP and solve times for experiments on the synthetic data are summarized in Figures~\ref{fig:athey_v1} and~\ref{fig:synthetic_comp_times}, respectively (more details can be found in Table~\ref{tab:synthetic_table}). For the MIO-based methods (ours, K-PT, B-PT) and PT, we train trees of depth 1 because \revision{one split is sufficient given our data generation process} (see equation~\eqref{eq:synthetic}). All methods solve to all instances in less than 9 seconds.
In the following, we omit our analyses on CF and CT for the synthetic data for brevity; refer to our analysis on the warfarin dataset for an in-depth discussion on the advantages and drawbacks for CF/CT.

\paragraph{Out-of-sample Performance.}
Figure~\ref{fig:athey_v1} shows the average OOSP for each method in dependence of the probability of correct treatment assignment, $p$. From the left subfigure, it can be seen that the DR and DM methods have competing performance at around 75\% OOSP across $p$, while IPW performs consistently worse ranging from 60-72\% OOSP. DM's superior performance aligns with our expectations since the linear regressor is an accurate predictor of counterfactual outcomes, and consequently DR benefits from this modeling choice as well.

Further, we observe that PT maintains a similar performance to our DR method as expected, since both methods optimize the same objective. Recall that PT learns from the raw, continuous features while DR takes in the discretized versions of these features; the competing performance between these methods indicates that, in this application, there is virtually no loss in performance when learning from discretized data.

We also empirically test DR's robustness property as discussed in Section~\ref{section:robust_intro}. The dotted and dashed lines on the Figure~\ref{fig:athey_v1} (right) represent instances when one of the two estimators in DR is inaccurate (i.e., uses Lasso to estimate $\mu$ or Log to estimate $\nu$). In these experiments, DR favors the other (better) estimator as expected, resulting in a relatively consistent performance across $p$ despite the suboptimal predictors.

These results directly contrast K-PT and B-PT\revision{, which are not consistent across $p$. Both methods obtain decent performances when $p=0.5$ at 70\% OOSP, i.e., when the historical policy is randomized -- this behavior is expected given their methods rely on Assumption~\ref{asu:sufficient_partition}. However}, both methods drop to around 50\% (i.e., similar to random treatment assignment) when $p \in \{0.75, 0.9\}$.} Their poor performance in these settings is particularly relevant since most observational data is based on an informed policy that presumably treats most patients correctly. The likely explanation for this drop in performance is that a tree of moderate depth is unlikely to partition the dataset sufficiently fine for Assumption~\ref{asu:sufficient_partition} to hold. 

\newnj{Finally, we plot R\&C (using linear regression) on Figure~\ref{fig:athey_v1} (left), which corresponds to the best performance of any predictive model (since our counterfactuals are constructed via a linear function, see equation~\eqref{eq:synthetic}). The difference in performance between R\&C and our trees quantifies the ``price of interpretability'', i.e., the trade-off in performance we observe when opting for a more interpretable model compared to a higher-performing, less interpretable model. Following \cite{jo2022learning}, we measure interpretability via a variant of ``decision complexity'': the minimum number of parameters required for a model to determine a given datapoint's treatment assignment. With this definition, our trees of depth 1 have a decision complexity of 3 (1 branching node and 2 leaf nodes), while the R\&C model has a complexity of 6: 2 linear models corresponding to the 2 treatment options, each having 2 coefficients and 1 bias term. The price of interpretability is a small gap of on average 2 percentage points (p.p.) in optimal treatment assignment.}

\begin{figure}[h]
\includegraphics[width=\textwidth]{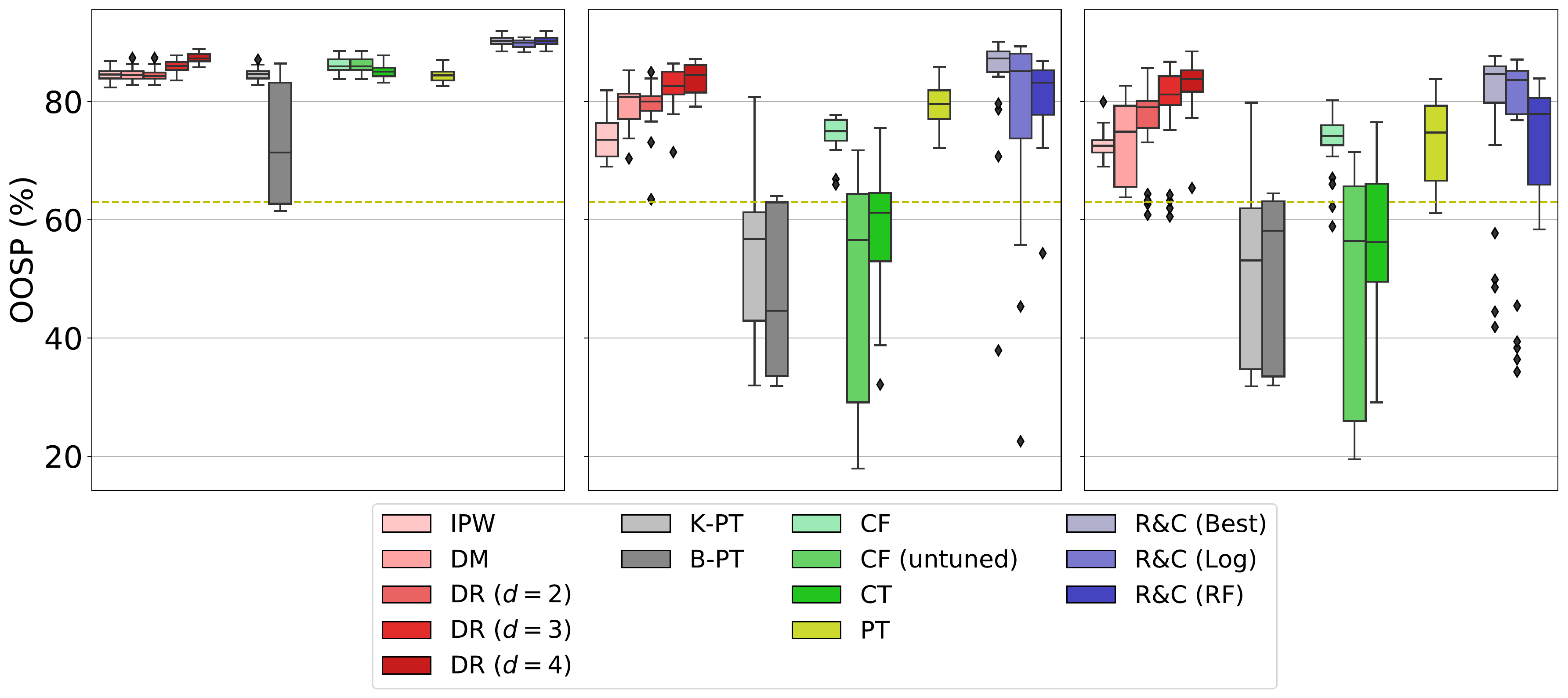}
\centering
\caption{\newnj{Out-of-sample probability of correct treatment assignment (OOSP) over all 5 realizations on the warfarin dataset. The left graph shows the distribution for the randomized experiments. The middle (resp.\ right) figure shows the distribution of results for $r = 0.06$ (resp.\ $0.11$). Unless mentioned otherwise, all MIO-based methods (IPW, DM, DR, K-PT, B-PT) and PT are averages over trees with depth $d=2$ with 4 leaf nodes.}}
\label{fig:warfarin_box}
\end{figure}

\begin{figure}[h]
\includegraphics[width=0.9\textwidth]{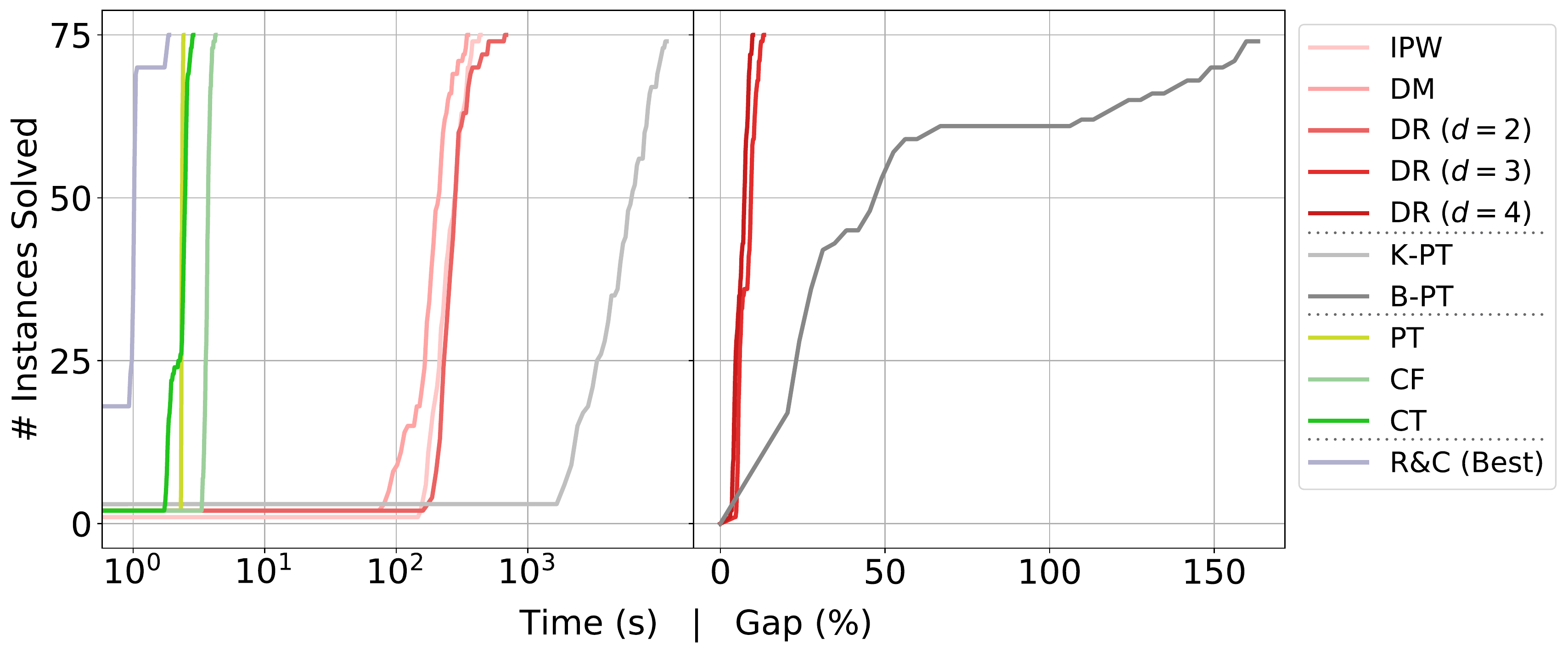}
\centering
\caption{\newnj{Comparison of computational performance for the warfarin data. All instances using the methods DR ($d=\{3, 4\}$) and B-PT did not solve to optimality within the 4-hour time limit, resulting in optimality gaps. All other methods solved to optimality within the time limit. We exclude CF (untuned) and R\&C (Log, RF) for brevity because the computational times are similar to CF and R\&C (Best), respectively.}}
\label{fig:warfarin_comp_times}
\end{figure}

\newnj{
\subsubsection{Warfarin Dosing.}\label{sec:exp_results_warfarin}

\paragraph{Optimization Performance.} The OOSP and computational times for our experiments on the warfarin data are summarized in Figures~\ref{fig:warfarin_box} and~\ref{fig:warfarin_comp_times}, respectively (more details can be found in Table~\ref{table:warfarin_table}). Unless otherwise noted, results for all MIO-based methods (ours, K-PT, B-PT) and PT are from trees of depth $d=2$. We chose this depth because \textit{1)} splitting on three features yield high enough performance, and \textit{2)} K-PT and B-PT did not scale to deeper trees and resulted in large optimality gaps, making for an unfair comparison. In the following, we also display results only for DR ($d \in \{3, 4\}$) in order to analyze the performance improvements on deeper optimal trees. 

In Figure~\ref{fig:warfarin_comp_times}, we see that the non-MIO methods find solutions consistently in less than 10 seconds. In contrast, our methods (for trees of depth 2) solve to optimality in the order of $10^2$ seconds, and K-PT in $10^3$ seconds. Our methods for deeper trees ($d \in \{3, 4\}$) do not solve to optimality within the 4-hour time limit, but have small optimality gaps (less than 20\%). Meanwhile, B-PT suffers from extremely large optimality gaps of $> 150\%$ due to the quadratic objective. 

\paragraph{Out-of-sample Performance.} We first compare our methods' performance across the three experimental designs. In the marginally randomized setting (Figure~\ref{fig:warfarin_box}, left), IPW, DM, and DR all achieve similar performance of on average 84.5\% OOSP. In the conditionally randomized settings ($r \in \{0.06, 0.11\}$, middle and right subfigures), DR performs at a little less than 80\% OOSP, while DM and IPW perform marginally worse average at 76\% and 73\% on average, respectively. This overall decrease in performance is expected given that the conditionally randomized settings are harder to solve, but the fact that DR maintains decent performance compared to DM and IPW indicates that this is generally the strongest method to use, as discussed in Section~\ref{section:robust_intro}.

When we increase the depth of our trees (DR, $d \in \{3, 4\}$), performance noticeably increases at on average 86.0\% and 87.3\% OOSP, respectively, despite all of the instances not solving to optimality within the time limit. Increasing depth, however, has diminishing returns -- ultimately, increasing a tree's depth introduces a trade-off between interpretability and marginal increases in performance that only practitioners can make in their respective domains.

Similar to the synthetic experiments, we observe that K-PT performs well in the marginally randomized setting with competing performance to our methods, while B-PT suffered from optimality gaps. However, in the conditionally randomized settings, both K-PT and B-PT not only have high variance, but also perform consistently worse than the simple predictor that only predicts $K=0$ for the entire population (yellow dashed line). While the simple predictor averages at around 63\% OOSP, K-PT and B-PT perform at on average 51\% and 48\% respectively. We note that all instances of K-PT solved to optimality, indicating that its objective is ill-suited for conditionally randomized experiments, as analyzed in Section~\ref{section:kallus}. B-PT, on the other hand, suffered from optimality gaps, but we can assume that its optimal performance might only be slightly better than K-PT since it suffers from similar problems as outlined in Section~\ref{section:bertsimas}. 

With regard to the CT and CF, both have similar performances to ours in the marginally randomized setting. CF, being a more complex method, has a slightly higher OOSP at 86.2\%. However, in the conditionally randomized settings, both methods have similar performances to K-PT and B-PT when their hyperparameters are untuned. As is the case with many black-box models, it becomes necessary to tune their hyperparameters so as to avoid overfitting and to tailor to the data. Indeed, when we tune CF, we see drastic performance increases at 74\% on average. In theory, CF could perform better than our DR method if more hyperparameters were searched since its model complexity is much higher. This process of hyperparameter tuning, however, illustrates a key trade-off between training a black-box causal model like CF -- which not only requires tuning but also an analysis on heterogeneity ex-post -- versus training an interpretable model (ours) -- which requires careful inference of counterfactual predictions and propensity weighting but is in itself transparent. \revisio{Moreover, we emphasize that CT and CF are designed to estimate heterogeneous treatment effects and not learn prescriptive policies. While we have adapted their method to do the latter, both approaches can only handle two treatment options, and thus when presented with more than two options, they can only make pairwise (local) decisions. In contrast, methods that are designed to prescribe policies like ours have a global view of treatment assignment. In the presence of more than two treatment options, CT -- despite it being a simple decision tree -- also becomes much less interpretable because it may produce partitions that differ from one treatment to another, and thus when comparing these treatments to build a prescriptive policy, CT ends up yielding a more complicated decision rule.}

We also compare our DR method to PT, which again optimizes the same objective except that PT can learn from continuous data. Similar to our observations for the synthetic experiments, PT maintains roughly the same performance as DR -- with the exception of the experiments when $r=0.11$. Here, we observe that PT has a higher variance and performs marginally worse at 73\% OOSP compared to 76\% for DR. We posit that this degradation in performance is a problem of overfitting when PT is given the full suite of continuous features. On the training set, PT performs similarly to DR (see Figure~\ref{fig:warfarin_overfitting} in Appendix~\ref{sec:pt_overfitting}), but in Figure~\ref{fig:warfarin_box} its trees fail to generalize as well out-of-sample. As such, discretizing continuous variables has the benefit of occasionally serving as a form of regularization to avoid overfitting.

Finally, we quantify the ``price of interpretability'' similar to the synthetic experiments when we compare our method (say, DR $d=2$) with the R\&C (Best) approach. In this case, DR's decision complexity is 7 (3 branching and 4 leaf nodes) while R\&C's decision complexity is in the order of $10^5$ (an exact number is hard to calculate given they are averages over a combination of random forests and logistic regression models, but for each random forest we train $10^2$ trees, each with around $10^3$ complexity). In the randomized setting and when $r=0.06$ (middle subfigure), R\&C performs on average 5.75 p.p. better than DR ($d=2$). That gap shrinks to 0.77 p.p. when $r=0.11$ (right subfigure). This price of interpretability is even smaller when we compare DR ($d=4$) and R\&C (Best); R\&C performs on average 2.9, 0.0, and 0.0 p.p. better than DR ($d=4$) on the randomized, $r=0.06$, and $r=0.11$ settings, respectively.}

\newnj{
\subsection{Experiments with Budget and Fairness Constraints}

As discussed in Section~\ref{sec:extensions}, our method has flexible modeling power; the MIO implementation allows for the addition of various constraints, which other methods like PT and CF cannot incorporate because they are learned recursively or via a heuristic. In this section, we perform additional experiments on our method that consider budget constraints on synthetic data and fairness constraints in the Warfarin dosing setting.

\subsubsection*{Budget Constraints on Synthetic Data.}
Equation~\ref{eq:synthetic} -- which models the potential outcomes as functions of covariates $X = (X^1, X^2)$ -- controls treatment assignment to only be affected by~$X^1$. In particular, the optimal split is easily modeled by a tree of depth 1 where observations with $X^1 > 0$ (resp. $X^1 \leq 0$) will be assigned treatment $K=1$ (resp. $K=0$), i.e., approximately half of the population receives either treatment. When we impose a constraint such that $K=1$ can only be given to $< 50\%$ of the population, the optimal split will shift towards larger values of $X^1$. We model such behavior in this set of experiments. We run the DR method similar to the experiments described in Section~\ref{sec:experimental_setup}, with the addition of a constraint
$\sum_{n \in \sets B \cup \sets T} \sum_{x \in \sets X} \sum_{i \in \sets{I}: X_i = x} z^{n,t_{1}}_x \; \leq \; |\sets{I}|C, $
with $C \in [0.05, 0.4]$ in 0.01 increments.

Figure~\ref{fig:synthetic_budget} shows OOSP for the doubly robust method in dependence of $C$, using either linear or lasso regression to estimate $\nu$ and either a decision tree or logistic regression to estimate $\mu$. Over all choices of estimators, there is a clear trend whereby tightening the budget on $K=1$ decreases performance; our unconstrained performance is as high as 78\% but goes down to 57\% OOSP when $C = 0.05$. This trend is expected as the policy can only assign $K=1$ to fewer people when on average around half of the population needs said treatment. The effect of using different estimators is also stark, particularly in settings where the historical policy is not randomized ($p \neq 0.5$). \revision{The experiments using LR and DT (top left of Figure~\ref{fig:synthetic_budget}) reflect the best-performing estimators. Indeed, similar to the results in Section~\ref{sec:exp_results_synthetic}, we see that using LR/DT leads to consistent performance across all experimental designs. In contrast, when we use a worse model for $\hat{\nu}$ (Lasso, right column in Figure~\ref{fig:synthetic_budget}) and/or a worse $\hat{\mu}$ (Log, bottom row in Figure~\ref{fig:synthetic_budget}), we observe that performance is degraded when $p \neq 0.5$.} As is the case with Figure~\ref{fig:athey_v1} (right), we see that the doubly robust estimator somewhat maintains a high performance when either one of the estimators ($\hat{\nu}$ or $\hat{\mu}$) is correct.

\begin{figure}[h]
\includegraphics[width=0.9\textwidth]{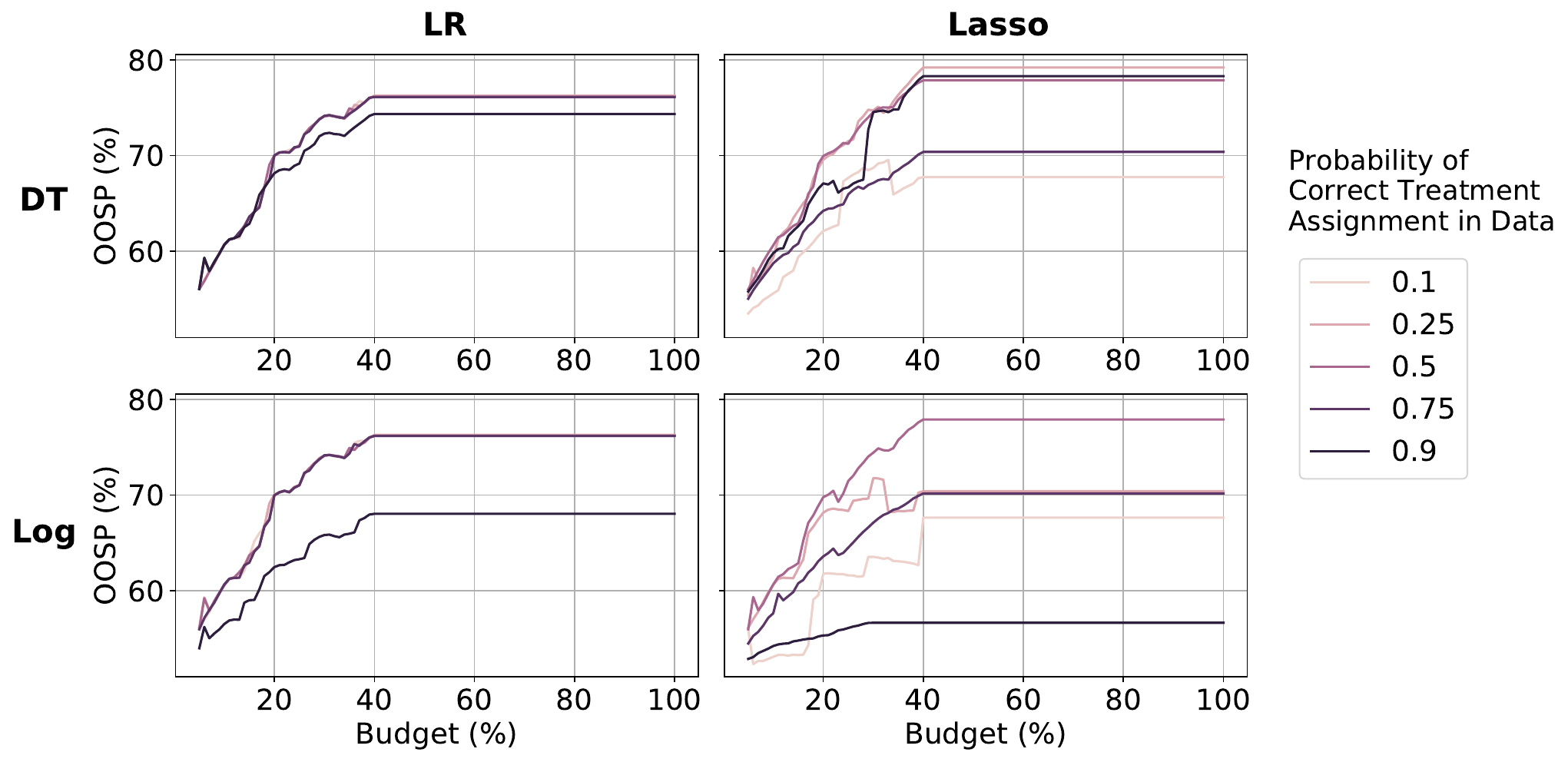}
\centering
\caption{\newnj{Out-of-sample probability of correct treatment assignment (OOSP) (y-axis) for the doubly robust method when the available budget -- \% of the population that can receive treatment $K=1$ -- varies (x-axis) and using different estimators for $\hat{\mu}$ and $\hat{\nu}$. Different line colors correspond to different probabilities of correct treatment assignment in the historical policy, $p$.}}
\label{fig:synthetic_budget}
\end{figure}

\subsubsection*{Fairness Constraints on Warfarin Dosing.}\label{sec:additional_experiments}
The optimal trees learned in Section~\ref{section:experimental_results} \revision{yield the worst outcomes for White patients}: over all trees, the median disparity in realized outcomes between White and non-White patients is 0.04 and 0.08 for the randomized and non-randomized settings, respectively, with no observations achieving full parity or the reverse disparity (being biased toward non-White people). Since our outcomes correspond to correct treatment assignment, this disparity can be interpreted as 4 and 8 percentage points fewer White people who receive their optimal treatment compared to non-White people. To close this gap, we run additional experiments imposing that the expected outcomes (as estimated by the DM objective) between White and non-White people be at most $\delta$, i.e.,
\begin{equation*}
\begin{split} \left|
M_{\text{white}} - M_{\text{!white}} \right|\; \leq \; \delta,
\end{split}
\end{equation*}
where 
$$M_p = \frac{1}{|\{i \in \sets{I}: P_i = p\}|} \displaystyle \sum_{n \in \sets B \cup \sets T} \displaystyle \sum_{k \in \sets{K}} \displaystyle \sum_{x \in \sets X : P_x = p}  z^{n,t_{k}}_{x} \displaystyle \sum_{i \in \sets I: X_i = x} \hat{\nu}_k(X_i).$$


We note that while the DR objective leads to generally better performance, we found that the DM method estimates outcomes that are closest to the true expected outcomes; we believe the DR method is less-suited to reflect true expected outcomes because the conditionally randomized settings have many observations with extremely low propensity weights, which severely biases the estimates. We let $\delta \in [0.01, 0.08]$ in 0.01 increments, and display realized disparity between the two groups as~$\delta$ varies in Figure~\ref{fig:warfarin_fairness} across the three experimental designs. As expected, decreasing $\delta$ has a direct relationship with a decrease in realized outcome disparity. However, we note that enforcing $\delta$ does not guarantee that actual disparity is within $\delta$ because we can only expect the constraint to hold in expectation rather than almost surely in a finite sample. For the randomized setting (left subfigure), enforcing $\delta = 0.01$ leads to a median disparity of around $-0.02$ across all seeds. Similarly, for the conditionally randomized settings, enforcing $\delta = 0.01$ and 0.02 leads to a median of full parity for $r=0.06$ and $r=0.11$, respectively.

\begin{figure}[h]
\includegraphics[width=\textwidth]{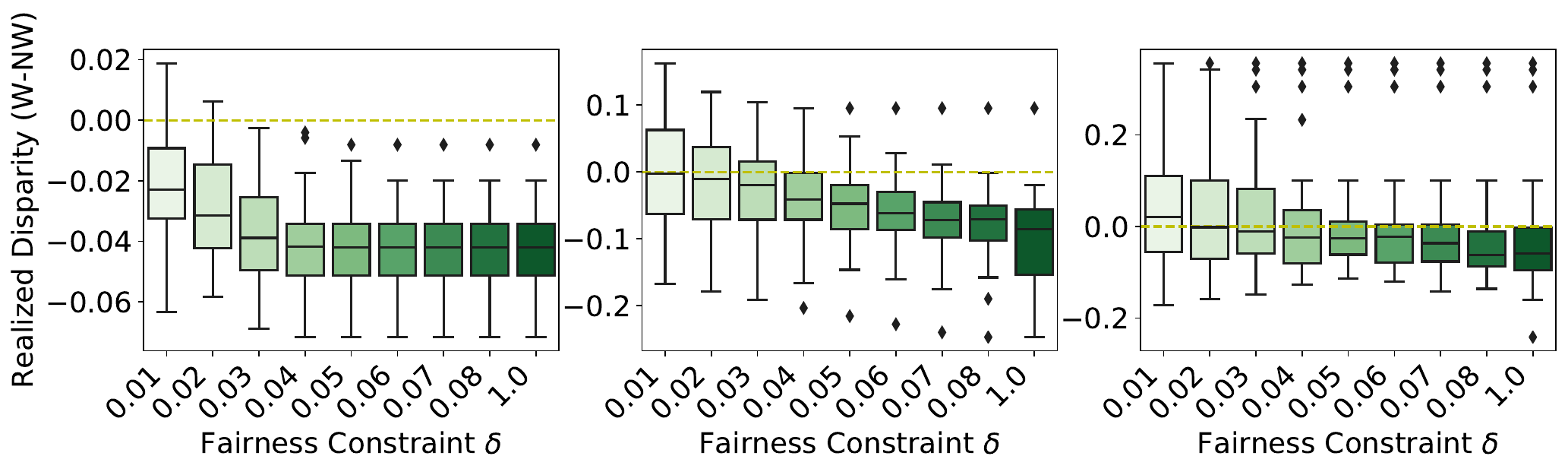}
\centering
\caption{\newnj{Disparity of realized outcomes between White (W) and non-White (NW) patients (y-axis) as fairness parameter $\delta$ varies (x-axis), over the randomized setting (left) and non-randomized settings ($r \in \{0.06, 0.11\}$ in the middle and right, respectively). The dashed yellow line indicates true parity.}}
\label{fig:warfarin_fairness}
\end{figure}

}

\section{Summary}
We presented MIO formulations to learn treatment assignment policies in the form of prescriptive trees from observational data. We showed that our methods are asymptotically exact, which sets us apart from existing literature on prescriptive trees that \newnj{\textit{1)} require data to originate from marginally randomized experiments; \textit{2)} require the learned trees to be very deep to yield correct treatment assignments; or \textit{3)} cannot handle budget and/or fairness constraints}. Our experiments show that our methods perform consistently better across different experiment designs, in some cases performing~$3 \times$ better than the state-of-the-art. \newnj{We also showcase our method's flexible modeling power by running additional experiments incorporating budget and fairness constraints: a feature that no other method in the literature can handle efficiently.}

\ACKNOWLEDGMENT{N.\ Jo acknowledges support from the Epstein Institute at the University of Southern California. P.\ Vayanos and S.\ Aghaei are funded in part by the National Science Foundation under CAREER award number 2046230. They are grateful for this support. N.\ Jo, P.\ Vayanos, and S.\ Aghaei gratefully acknowledge support from the Hilton C.\ Foundation, the Homeless Policy Research Institute, and the Home for Good foundation under the ``C.E.S.\ Triage Tool Research \& Refinement'' grant. A.\ G\'omez is funded in part by the National Science Foundation under grant 2006762.}

\bibliographystyle{informs2014}


\bibliography{bib.bib}

\newpage





\ECSwitch


\ECHead{Electronic Companion}

\section{K-PT MIO Formulation}
\label{section:kallus_formulation}

In this section, we provide an adapted version of the formulation from \cite{Kallus2017RecursiveData} that we use in our experiments in Section~\ref{section:experimental_results}. As we will soon discuss, we adapt the original formulation for both brevity and so that it better aligns with the notation we use. Extending the notation from the problem statement, we construct a perfect binary tree and number the nodes 1 through $2^{d+1}-1$ in the order in which they appear on a breadth-first-first search. We let $R_{nm} \in \{1, -1\}$ capture the relationship of node $n$ with its ancestor(s) $m \in \sets A(n)$, i.e., $R_{nm}$ equals $1$ iff we use $m$'s right branch to reach $n$, and $-1$ otherwise. 

We now define the decision variables used in the K-PT formulation. For every branching node $n\in \sets B$ and feature $f\in \sets F$, we let the binary variable $b_{nf}$ indicate if feature~$f$ is selected for branching at node $n$. We let the binary variable $\chi_{in}$ equal 1 iff datapoint $i$ goes left on branching node $n$. Further, we define membership variables $\lambda_{in} \in \{0, 1\}$, which takes value 1 iff datapoint $i$ flows to terminal node $n \in \sets{T}$. For $k\in \sets K$, we let $w_{nk}\in \{0,1\}$ equal~1 iff treatment $k$ is selected at terminal node $n$. Also for each terminal node $n$, define $\rho_{n}$ to be the average treatment outcome of all datapoints in that node. Let $\eta_{in}$ be the product of $\lambda_{in}$ and $\rho_{n}$, which captures the average outcome associated with datapoint $i$. 

In order to linearize $\eta$, \cite{Kallus2017RecursiveData} introduces big-$M$ constraints. To this end, let $\overline{Y}_i = Y_i - \min_{j \in \mathcal I} Y_j$, $\overline{Y}_{max} = \max_{i}\overline{Y}_i$, and $M = \overline{Y}_{max} (\max_{k \in K} \sum_{i \in \mathcal I} \mathbb{I}[K_i = k])$. The approach now reads:
\begin{subequations}
\begin{align}
\text{maximize} \;\; & \displaystyle \sum_{i \in \mathcal I} \sum_{n \in \sets T } \eta_{in} 
\label{eq:kallus_obj}\\
\text{subject to}\;\;
& \displaystyle \lambda_{in} \leq \frac{1+R_{nm}}{2} - R_{nm} \chi_{im}   &\hspace{-5cm}  \forall i \in \mathcal I, n \in \sets T, m \in a(n) \label{eq:kallus_membership1}\\
&  \displaystyle \lambda_{in} \geq 1-\displaystyle \sum_{\substack{m \in a(n) \\ R_{mn} = 1}} \chi_{im} + \displaystyle \sum_{\substack{m \in a(n) \\ R_{mn} = -1}} (-1+\chi_{im})  &  \forall i \in \mathcal I, n \in \sets T \label{eq:kallus_membership2}\\
& \displaystyle \sum_{f \in \mathcal F} b_{nf} = 1  &\hspace{-5cm}  \forall n \in \sets B \label{eq:kallus_branching_constraint}\\
&  \displaystyle \chi_{in} = \sum_{\substack{f \in \mathcal F: \\ X^i_f = 0}} b_{nf} &\hspace{-5cm}   \forall i \in \mathcal I,n \in \sets B \label{eq:kallus_branching_chi}\\
\comment{& \displaystyle \sum_{i: K_i = k} \lambda_{in} \geq n_{\text{min-leaf}} &\hspace{-5cm} \forall k \in \mathcal K, n \in \sets T \label{eq:kallus_minleaf}\\}
& \eta_{in} \leq \overline{Y}_{max} \lambda_{in}, \eta_{in} \leq \rho_{n} & \forall i \in \mathcal I, p \in \sets T \label{eq:kallus_nu_linearize1}\\
& \eta_{in} \geq \rho_{n}-\overline{Y}_{max} (1-\lambda_{in}) & \forall i \in \mathcal I, p \in \sets T \label{eq:kallus_nu_linearize2}\\
& \displaystyle\sum\limits_{i:K_i = k} (\eta_{in}-\lambda_{in}\overline{Y}_{i}) \leq M (1-w_{nk}) & \forall p \in \sets T, k \in \sets K \label{eq:kallus_M1}\\
& \displaystyle\sum\limits_{i:K_i = k} (\eta_{in}-\lambda_{in}\overline{Y}_{i}) \geq M (w_{nk}-1) & \forall p \in \sets T, k \in \sets K \label{eq:kallus_M2}\\
&  \displaystyle \sum_{k \in K} w_{nk} = 1 &\hspace{-5cm} \forall n \in \sets T \label{eq:kallus_treatment_constraint}\\
&  \displaystyle w_{nk} \in \{0,1\}  &\hspace{-5cm}   \forall n \in \sets T,k \in \sets K \\
&  \displaystyle b_{nf} \in \{0,1\}  &\hspace{-5cm}   \forall n \in \sets B,f \in \sets F \\
&  \displaystyle \lambda_{in} \in \{0,1\}  &\hspace{-5cm}   \forall i \in \mathcal I, n \in \sets T \\
&  \displaystyle \chi_{in} \in \{0,1\}  &\hspace{-5cm}  \forall n \in \sets B,i \in \mathcal I\\
&  \displaystyle \eta_{in} \in \mathbb{R}_+  &\hspace{-5cm}  \forall n \in \sets T,i \in \mathcal I\\
&  \displaystyle \rho_{n} \in \mathbb{R}_+  &\hspace{-5cm}  \forall n \in \sets T.
\end{align}
\label{eq:kallus}
\end{subequations}

The objective function \eqref{eq:kallus_obj} sums the associated averages over all datapoints and terminal nodes. Constraints~\eqref{eq:kallus_membership1} and~\eqref{eq:kallus_membership2} mode the flow of datapoints from the root down to the terminal nodes. Constraints~\eqref{eq:kallus_branching_constraint} state that, at each branching node, the tree must branch on exactly one feature, whereas~\eqref{eq:kallus_branching_chi} defines~$\chi$ using the associated branching decisions. Constraints~\eqref{eq:kallus_nu_linearize1} and~\eqref{eq:kallus_nu_linearize2} both use big-$M$ constraints to linearize~$\eta$. Constraints ~\eqref{eq:kallus_M1} and~\eqref{eq:kallus_M2} ensure that the predicted outcome for a given treatment at each leaf of the tree is indeed the average observed outcome. Lastly, constraint \eqref{eq:kallus_treatment_constraint} ensures that exactly one treatment is assigned at every terminal node. 

Note that we have here converted the formulation from \cite{Kallus2017RecursiveData} to a maximization problem to account for the different representation of outcome~$Y$ (higher values preferred). Further, we make the following changes from the formulation presented in \cite{Kallus2017RecursiveData}: \textit{(i)} constraint \eqref{eq:kallus_membership2} is slightly different, since the original constraint does not result in a correct behavior; \textit{(ii)} for simplicity, we removed constraints inspired from \cite{yildiz2013incremental} from the formulation and replaced it with constraint \eqref{eq:kallus_branching_constraint}, which serves the same function;  \textit{(iii)} we removed constraint $\sum_{i:K_i=k}\lambda_{in}\geq N_\text{min}$ from the formulation to allow for an equivalent comparison with our proposed methods, and in tuning the parameter for our experiments, we did not find statistical improvements.

\delag{constraint \eqref{eq:kallus_membership2} is different above compared to~\cite{Kallus2017RecursiveData} to account for an error in the formulation. \todiscusspv{not sure that we want to emphasize this? let's ask Andres what he thinks} For clarity, constraint~\eqref{eq:kallus_branching_constraint} replaces a constraint from the original formulation that serves the same purpose. There was also a constraint in the original formulation controlling how many instances from each treatment group must be in each terminal node. In testing, however, we found no significant improvement in adjusting this variable; we decided to omit the constraint to allow for an equivalent comparison with our proposed methods.}

\vspace{5mm}
\section{B-PT MIO Formulation}
\label{section:bertsimas_formulation}
\newnj{\cite{Bertsimas2019OptimalTrees} briefly discuss a coordinate descent algorithm to learn an optimal decision tree with the loss function form
$$\min_{T} \text{error}(T, D) + \alpha \cdot \text{complexity}(T),$$
where $T$ is the tree being optimized, $D$ is the training data, the ``error'' function measures how well the tree fits training data, and the second term is a penalization term that controls the trade-off between the quality of the tree and the tree's complexity/size. The algorithm \revision{iterates through different branching decisions and treatment assignment options until no possible improcements are found (i.e., the tree is a local minimum).} This process is repeated for different randomly generated starting trees, and ``the lowest objective function is selected as the final solution''. The documentation for this method is limited and there is no existing codebase allowing us to replicate this algorithm. Hence, in this section, we describe the MIO implementation that we use in our experiments for B-PT, which is adapted from~\cite{Kallus2017RecursiveData} to optimize the objective function proposed by \cite{Bertsimas2019OptimalTrees}.}

Building from \cite{Kallus2017RecursiveData}, \cite{Bertsimas2019OptimalTrees} added a regularization term that penalizes trees whose leaves have high variance. Let~$\beta_{nk}$ be the average outcome over all datapoints that were assigned treatment~$k$ in terminal node~$n$. Let~$g_i$ be the \emph{empirical} average of the outcomes at the terminal node where datapoint~$i$ lands over all datapoints that were assigned the same treatment as~$i$ (i.e. $g_i = \beta_{nK_{i}}$). Finally, let $\theta$ control the regularization strength. The formulation becomes:
\begin{subequations}
\begin{align}
\text{maximize} \;\; & \theta \displaystyle\sum\limits_{i \in \mathcal I} \displaystyle\sum\limits_{n \in T} \eta_{in} - (1-\theta) \displaystyle\sum\limits_{i \in \mathcal I} (Y_i - g_i)^2
\label{eq:bertsimas_obj}\\
\text{subject to}\;\;
& \displaystyle \lambda_{in} \leq \frac{1+R_{nm}}{2} - R_{nm} \chi_{im}   &\hspace{-5cm}  \forall i \in \mathcal I, n \in \sets T, m \in a(n) \label{eq:bertsimas_membership1}\\
&  \displaystyle \lambda_{in} \geq 1-\displaystyle \sum_{\substack{m \in a(n) \\ R_{mn} = 1}} \chi_{im} + \displaystyle \sum_{\substack{m \in a(n) \\ R_{mn} = -1}} (-1+\chi_{im})  &  \forall i \in \mathcal I, n \in \sets T \label{eq:berstimas_membership2}\\
& \displaystyle \sum_{f \in \mathcal F} b_{nf} = 1  &\hspace{-5cm}  \forall n \in \sets B \label{eq:bertsimas_branching_constraint}\\
&  \displaystyle \chi_{in} = \sum_{\substack{f \in \mathcal F: \\ X^i_f = 0}} b_{nf} &\hspace{-5cm}   \forall i \in \mathcal I,n \in \sets B \label{eq:bertsimas_branching_chi}\\
\comment{& \displaystyle \sum_{i: K_i = k} \lambda_{in} \geq n_{\text{min-leaf}} &\hspace{-5cm} \forall k \in \mathcal K, n \in \sets T \label{eq:bertsimas_minleaf}\\}
& \eta_{in} \leq \overline{Y}_{max} \lambda_{in}, \eta_{in} \leq \rho_{n} & \forall i \in \mathcal I, p \in \sets T \label{eq:bertsimas_nu_linearize1}\\
& \eta_{in} \geq \rho_{n}-\overline{Y}_{max} (1-\lambda_{in}) & \forall i \in \mathcal I, p \in \sets T \label{eq:bertsimas_nu_linearize2}\\
& \displaystyle\sum\limits_{i:K_i = k} (\eta_{in}-\lambda_{in}\overline{Y}_{i}) \leq M (1-w_{nk}) & \forall p \in T, k \in \sets K \label{eq:bertsimas_M1}\\
& \displaystyle\sum\limits_{i:K_i = k} (\eta_{in}-\lambda_{in}\overline{Y}_{i}) \geq M (w_{nk}-1) & \forall p \in T, k \in \sets K \label{eq:bertsimas_M2}\\
&  \displaystyle \sum_{k \in K} w_{nk} = 1 &\hspace{-5cm} \forall n \in \sets T \label{eq:bertsimas_treatment_constraint}\\
& g_i - \beta_{nK_i} \leq M (1-\lambda_{in}) & \forall i \in \mathcal I, n \in \sets T \label{eq:bertsimas_beta1} \\
& g_i - \beta_{nK_i} \geq M (\lambda_{in}-1) & \forall i \in \mathcal I, n \in \sets T \label{eq:bertsimas_beta2}\\
&  \displaystyle w_{nk} \in \{0,1\}  &\hspace{-5cm}   \forall n \in \sets T,k \in \sets K \\
&  \displaystyle b_{nf} \in \{0,1\}  &\hspace{-5cm}   \forall n \in \sets B,f \in \sets F \\
&  \displaystyle \lambda_{in} \in \{0,1\}  &\hspace{-5cm}   \forall i \in \mathcal I, n \in \sets T \\
&  \displaystyle \chi_{in} \in \{0,1\}  &\hspace{-5cm}  \forall n \in \sets B,i \in \mathcal I\\
&  \displaystyle \eta_{in} \in \mathbb{R}_+  &\hspace{-5cm}  \forall n \in \sets T,i \in \mathcal I\\
&  \displaystyle \rho_{n} \in \mathbb{R}_+  &\hspace{-5cm}  \forall n \in \sets T\\
&  \displaystyle g_{i} \in \mathbb{R}_+  &\hspace{-5cm}  \forall n \in \sets T.
\end{align}
\label{eq:bertsimas}
\end{subequations}
The objective function (\ref{eq:bertsimas_obj}) now penalizes high variance. All constraints remain the same with the exception of (\ref{eq:bertsimas_beta1}) and (\ref{eq:bertsimas_beta2}), which force~$g_i$ to be~$\beta_{nK_i}$ when datapoint~$i$ lands in terminal node~$n$. There are no additional constraints that define~$\beta_{nk}$ because the optimal solution for~$\min_{\beta_{nk}} \sum_{i=1}^m (a_i-\beta_{nk})^2$ is~$\beta_{nk}^* = \sum_{i=1}^m \frac{a_i}{m}$, where $\{a_1, \ldots, a_m\}$ are datapoints in terminal node~$n$ that were assigned treatment~$k$ in the data.

\vspace{5mm}
\section{Warfarin Dosage}
\label{sec:appendix_warfarin}

\vspace{3mm}

We present an equation from~\cite{international2009estimation} that we use in our experiments in Section~\ref{section:experimental_results} to determine the optimal warfarin dose for a patient. In particular, it allows us to generate patient counterfactuals and evaluate the learned policies for our experiments. Note that VKORC1 and CYP2C9 denote genotypes. If $W$ is the optimal weekly warfarin dosage, then the equation becomes:

\begin{equation}\label{warfarin_function}
\begin{split}
& \sqrt{W} = 5.6044 - 0.2614 \times \text{Age in decades} + 0.0087 \times \text{Height in cm} + 0.0128 \times \text{Weight in kg} \\ 
& - 0.8677 \times \text{VKORC1 A/G} - 1.6974 \times \text{VKORC1 A/A} - 0.4854 \times \text{VKORC1 genotype unknown} \\
& - 0.5211 \times \text{CYP2C9*1/*2}- 0.9357 \times \text{CYP2C9*1/*3} - 1.0616 \times \text{CYP2C9*2/*2} \\
& - 1.9206 \times \text{CYP2C9*2/*3} - 2.3312 \times \text{CYP2C9*3/*3} - 0.2188 \times \text{CYP2C9 genotype unknown} \\
& - 0.1092 \times \text{Asian race} - 0.2760 \times \text{Black or African American} - 0.1032 \times \text{Missing or Mixed race} \\
& + 1.1816 \times \text{Enzyme inducer status} - 0.5503 \times \text{Amiodarone status}
\end{split}
\end{equation}

\newpage
\section{Experiment Results (Raw)}
\label{section:experiments_detail}

\vspace{1mm}

In this section, we provide tables that contain the raw results for the experiments summarized in Section~\ref{section:experimental_results}. 
\newnj{
\begin{table}[H]
\caption{\revision{Companion table to the experiments on synthetic data in Section~\ref{sec:exp_results_synthetic}. The first and second number correspond to the average and standard deviation of the optimality gap, solving time, out-of-sample (OOS) regret, and OOS probability of correct treatment assignment (OOSP) across 5 random samples.}}
\tiny
\resizebox{\textwidth}{!}{
\begin{tabular}{@{}lllllll@{}}
\toprule
Depth & Method & Model       & Gap                  & Solving Time (s)           & OOS Regret             & OOSP (\%)              \\ 
\midrule
    1 &    IPW &         DT & 0.00 ± 0.00 & 1.34 ± 0.22 &  123.18 ± 95.63 &   66.25 ± 15.84 \\
    1 &    IPW &        Log & 0.00 ± 0.00 & 1.31 ± 0.31 & 138.12 ± 114.65 &   63.89 ± 18.00 \\
    1 &     DM &         LR & 0.00 ± 0.00 & 0.76 ± 0.40 &   64.22 ± 38.06 &    75.28 ± 9.87 \\
    1 &     DM &      Lasso & 0.00 ± 0.00 & 0.57 ± 0.09 &  200.68 ± 72.26 &   53.44 ± 11.00 \\
    1 &     DR &     DT, LR & 0.00 ± 0.00 & 0.86 ± 0.46 &   65.76 ± 40.68 &   75.01 ± 10.13 \\
    1 &     DR &  DT, Lasso & 0.00 ± 0.00 & 2.08 ± 0.82 &   77.04 ± 70.23 &   73.70 ± 12.83 \\
    1 &     DR &    Log, LR & 0.00 ± 0.00 & 0.95 ± 0.56 &   78.78 ± 55.88 &   72.98 ± 11.56 \\
    1 &     DR & Log, Lasso & 0.00 ± 0.00 & 1.79 ± 1.01 &  111.47 ± 68.44 &   67.16 ± 11.88 \\
    1 &   K-PT &          - & 0.00 ± 0.00 & 1.97 ± 0.76 & 161.83 ± 103.62 &   60.10 ± 17.46 \\
    1 &   B-PT &          - & 0.00 ± 0.00 & 6.28 ± 1.36 & 179.04 ± 106.95 &   57.62 ± 17.63 \\
    1 &     PT &     DT, LR & 0.00 ± 0.00 & 0.01 ± 0.00 &   62.01 ± 37.00 &   75.25 ± 10.29 \\
    - &     CF &          - & 0.00 ± 0.00 & 0.75 ± 0.04 &   76.03 ± 57.20 &   73.13 ± 12.05 \\
    - &     CT &          - & 0.00 ± 0.00 & 0.29 ± 0.02 &   92.20 ± 74.63 &   71.53 ± 13.64 \\
    - &    R\&C &         LR & 0.00 ± 0.00 & 0.23 ± 0.00 &   51.81 ± 32.87 &   77.11 ± 10.09 \\
\bottomrule
\end{tabular}}
    \label{tab:synthetic_table}
\end{table}}

\begin{table}[h]
\caption{\revision{Companion table to the experiments on synthetic data in Section~\ref{sec:exp_results_warfarin}. The first and second number correspond to the average and standard deviation of the optimality gap, solving time, out-of-sample (OOS) regret, and OOS probability of correct treatment assignment (OOSP) across 25 random samples.}}
\tiny
\resizebox{\textwidth}{!}{
\begin{tabular}{@{}lllllll@{}}
\toprule
Depth & Method & Model  & Gap             & Solving Time (s)         & OOS Regret               & OOSP (\%)              \\ \midrule
    2 &          IPW &         DT &   0.00 ± 0.00 &    255.45 ± 65.13 &  317.39 ± 82.90 &    77.10 ± 5.98 \\
    2 &           DM &     RF/Log &   0.00 ± 0.00 &    192.03 ± 64.13 &  288.79 ± 88.46 &    79.16 ± 6.38 \\
    2 &           DR & DT, RF/Log &   0.00 ± 0.00 &    284.58 ± 83.83 &  279.32 ± 87.00 &    79.85 ± 6.28 \\
    2 &         K-PT &          - &   0.00 ± 0.03 & 5435.72 ± 2792.71 & 522.12 ± 264.43 &   62.33 ± 19.08 \\
    2 &         B-PT &          - & 52.30 ± 43.78 & 14333.53 ± 674.39 & 601.39 ± 239.26 &   56.61 ± 17.26 \\
    2 &           PT &  DT, RF/Log &   0.00 ± 0.00 &       2.35 ± 0.03 &  257.73 ± 76.39 &    81.40 ± 5.51 \\
     \midrule
    3 &          IPW &         DT &  13.56 ± 4.15 &   14405.89 ± 0.50 &  246.35 ± 48.07 &    82.23 ± 3.47 \\
    3 &           DM &     RF/Log &   2.10 ± 1.05 &   14406.33 ± 0.55 &  252.92 ± 69.22 &    81.75 ± 4.99 \\
    3 &           DR & DT, RF/Log &   8.08 ± 2.38 &   14406.65 ± 0.70 &  243.00 ± 78.18 &    82.47 ± 5.64 \\
     \midrule
    4 &          IPW &         DT &   9.51 ± 2.91 &   14412.07 ± 1.17 &  221.00 ± 39.47 &    84.05 ± 2.85 \\
    4 &           DM &     RF/Log &   1.48 ± 0.73 &   14412.46 ± 1.05 &  243.23 ± 72.31 &    82.45 ± 5.22 \\
    4 &           DR & DT, RF/Log &   6.26 ± 1.90 &   14413.10 ± 1.57 &  210.95 ± 48.01 &    84.78 ± 3.46 \\
     \midrule
    - &           CF &          - &   0.00 ± 0.00 &       3.67 ± 0.19 &  304.47 ± 94.22 &    78.03 ± 6.80 \\
    - & CF (untuned) &          - &   0.00 ± 0.00 &       3.75 ± 0.33 & 542.52 ± 334.07 &   60.86 ± 24.10 \\
    - &           CT &          - &   0.00 ± 0.00 &       2.31 ± 0.32 & 459.35 ± 223.37 &   66.86 ± 16.12 \\
    - &          R\&C &       Best &   0.00 ± 0.00 &       0.95 ± 0.19 & 225.89 ± 163.98 &   83.70 ± 11.83 \\
    - &          R\&C &        Log &   0.00 ± 0.00 &       0.12 ± 0.01 & 261.47 ± 198.86 &   81.14 ± 14.35 \\
    - &          R\&C &         RF &   0.00 ± 0.00 &       0.83 ± 0.03 & 253.59 ± 125.24 &    81.70 ± 9.04 \\
\bottomrule
\end{tabular}}
\label{table:warfarin_table}
\end{table}



\begin{table}[h]
\caption{\revision{Mean and standard deviation statistics over 5 random samples in the synthetic experiments, with a constraint that treatment $K=1$ is available to at most $C$\% in the population.}}
\label{table:budget_constraints}
\tiny
\centering
\resizebox{\textwidth}{!}{
\begin{tabular}{@{}llllllll@{}}
\toprule
Depth & Method & Model  & Budget Range ($C$)  & Gap             & Solving Time (s)         & OOS Regret               & OOSP (\%)              \\ \midrule
          1 &     DR &     LR, DT & 0.05-0.09 & 0.00 ± 0.00 & 0.16 ± 0.03 & 158.32 ± 49.39 &          58.35 ± 9.03 \\
          1 &     DR &     LR, DT & 0.10-0.14 & 0.00 ± 0.00 & 0.22 ± 0.13 & 132.80 ± 43.61 &          61.54 ± 9.04 \\
          1 &     DR &     LR, DT & 0.15-0.19 & 0.00 ± 0.00 & 0.23 ± 0.10 & 113.91 ± 44.06 &          65.44 ± 9.48 \\
          1 &     DR &     LR, DT & 0.20-0.24 & 0.00 ± 0.00 & 0.20 ± 0.11 &  85.58 ± 32.88 &          70.01 ± 8.02 \\
          1 &     DR &     LR, DT & 0.25-0.29 & 0.00 ± 0.00 & 0.23 ± 0.13 &  77.34 ± 32.68 &          72.25 ± 8.12 \\
          1 &     DR &     LR, DT & 0.30-0.34 & 0.00 ± 0.00 & 0.23 ± 0.13 &  69.97 ± 34.71 &          73.71 ± 8.56 \\
          1 &     DR &     LR, DT & 0.35-0.40 & 0.00 ± 0.00 & 0.23 ± 0.14 &  66.32 ± 36.18 &          75.04 ± 8.86 \\
          \midrule
          1 &     DR &    LR, Log & 0.05-0.09 & 0.00 ± 0.00 & 0.19 ± 0.11 & 163.39 ± 46.47 &          57.67 ± 8.51 \\
          1 &     DR &    LR, Log & 0.10-0.14 & 0.00 ± 0.00 & 0.22 ± 0.15 & 139.03 ± 41.97 &          60.64 ± 8.44 \\
          1 &     DR &    LR, Log & 0.15-0.19 & 0.00 ± 0.00 & 0.23 ± 0.13 & 119.83 ± 45.13 &          64.53 ± 9.18 \\
          1 &     DR &    LR, Log & 0.20-0.24 & 0.00 ± 0.00 & 0.22 ± 0.16 &  93.20 ± 40.82 &          68.85 ± 8.22 \\
          1 &     DR &    LR, Log & 0.25-0.29 & 0.00 ± 0.00 & 0.24 ± 0.16 &  85.11 ± 42.53 &          70.99 ± 8.61 \\
          1 &     DR &    LR, Log & 0.30-0.34 & 0.00 ± 0.00 & 0.24 ± 0.16 &  77.97 ± 44.93 &          72.40 ± 9.07 \\
          1 &     DR &    LR, Log & 0.35-0.40 & 0.00 ± 0.00 & 0.26 ± 0.18 &  74.07 ± 46.99 &          73.80 ± 9.65 \\
          \midrule
          1 &     DR &  Lasso, DT & 0.05-0.09 & 0.00 ± 0.00 & 0.29 ± 0.13 & 172.73 ± 59.95 &         56.89 ± 10.02 \\
          1 &     DR &  Lasso, DT & 0.10-0.14 & 0.00 ± 0.00 & 0.33 ± 0.17 & 150.23 ± 63.71 &         60.15 ± 10.97 \\
          1 &     DR &  Lasso, DT & 0.15-0.19 & 0.00 ± 0.00 & 0.37 ± 0.19 & 129.98 ± 70.03 &         63.92 ± 12.14 \\
          1 &     DR &  Lasso, DT & 0.20-0.24 & 0.00 ± 0.00 & 0.46 ± 0.24 & 110.86 ± 71.28 &         67.15 ± 12.24 \\
          1 &     DR &  Lasso, DT & 0.25-0.29 & 0.00 ± 0.00 & 0.47 ± 0.27 &  98.02 ± 64.04 &         69.78 ± 11.28 \\
          1 &     DR &  Lasso, DT & 0.30-0.34 & 0.00 ± 0.00 & 0.46 ± 0.30 &  84.12 ± 60.11 &         72.13 ± 10.78 \\
          1 &     DR &  Lasso, DT & 0.35-0.40 & 0.00 ± 0.00 & 0.41 ± 0.23 &  79.02 ± 64.46 &         73.62 ± 11.83 \\
          \midrule
          1 &     DR & Lasso, Log & 0.05-0.09 & 0.00 ± 0.00 & 0.27 ± 0.14 & 180.83 ± 65.74 &         55.80 ± 10.47 \\
          1 &     DR & Lasso, Log & 0.10-0.14 & 0.00 ± 0.00 & 0.29 ± 0.16 & 164.75 ± 74.07 &         57.94 ± 11.64 \\
          1 &     DR & Lasso, Log & 0.15-0.19 & 0.00 ± 0.00 & 0.34 ± 0.18 & 148.86 ± 75.23 &         60.86 ± 12.30 \\
          1 &     DR & Lasso, Log & 0.20-0.24 & 0.00 ± 0.00 & 0.34 ± 0.20 & 130.36 ± 73.55 &         63.94 ± 12.10 \\
          1 &     DR & Lasso, Log & 0.25-0.29 & 0.00 ± 0.00 & 0.40 ± 0.25 & 124.98 ± 73.15 &         65.19 ± 12.03 \\
          1 &     DR & Lasso, Log & 0.30-0.34 & 0.00 ± 0.00 & 0.36 ± 0.23 & 117.06 ± 76.89 &         66.58 ± 12.74 \\
          1 &     DR & Lasso, Log & 0.35-0.40 & 0.00 ± 0.00 & 0.36 ± 0.22 & 114.78 ± 76.44 &         67.13 ± 12.96 \\
\bottomrule
\end{tabular}}
\end{table}

\begin{table}[h]
\caption{\revision{Mean and standard deviation statistics over 25 random samples in the warfarin experiments, with a constraint that the difference between estimated outcomes between White and non-White patients is at most $
\delta$.}}
\tiny
\centering
\resizebox{\textwidth}{!}{
\begin{tabular}{@{}llllllllll@{}}
\toprule
Depth & Method & Model  & Fairness ($\delta$) & Gap  & Solving Time (s) & Est. Outcome Disparity & Actual Disparity & OOS Regret  & OOSP (\%)              \\ \midrule
          2 &     DM & RF/Log &      0.01 & 0.00 ± 0.00 & 316.31 ± 112.04 & -0.01 ± 0.01 &        0.01 ± 0.10 & 326.02 ± 106.02 &          0.76 ± 0.08 \\
          2 &     DM & RF/Log &      0.02 & 0.00 ± 0.00 & 284.67 ± 125.32 & -0.02 ± 0.01 &       -0.00 ± 0.10 & 321.20 ± 107.86 &          0.77 ± 0.08 \\
          2 &     DM & RF/Log &      0.03 & 0.00 ± 0.00 & 290.46 ± 130.17 & -0.02 ± 0.02 &       -0.01 ± 0.10 & 312.77 ± 105.23 &          0.77 ± 0.08 \\
          2 &     DM & RF/Log &      0.04 & 0.00 ± 0.00 & 280.67 ± 116.49 & -0.03 ± 0.02 &       -0.02 ± 0.09 &  303.54 ± 99.35 &          0.78 ± 0.07 \\
          2 &     DM & RF/Log &      0.05 & 0.00 ± 0.00 & 271.45 ± 123.14 & -0.03 ± 0.03 &       -0.03 ± 0.09 &  297.82 ± 96.61 &          0.79 ± 0.07 \\
          2 &     DM & RF/Log &      0.06 & 0.00 ± 0.00 & 268.73 ± 151.18 & -0.03 ± 0.03 &       -0.03 ± 0.09 &  294.58 ± 93.16 &          0.79 ± 0.07 \\
          2 &     DM & RF/Log &      0.07 & 0.00 ± 0.00 & 241.31 ± 111.07 & -0.04 ± 0.04 &       -0.04 ± 0.09 &  290.84 ± 92.17 &          0.79 ± 0.07 \\
          2 &     DM & RF/Log &      0.08 & 0.00 ± 0.00 & 234.19 ± 105.39 & -0.04 ± 0.04 &       -0.04 ± 0.09 &  290.78 ± 92.97 &          0.79 ± 0.07 \\
\bottomrule
\end{tabular}}
\label{table:fairness_constraints}
\end{table}

\begin{figure}[h]
\includegraphics[width=0.7\textwidth]{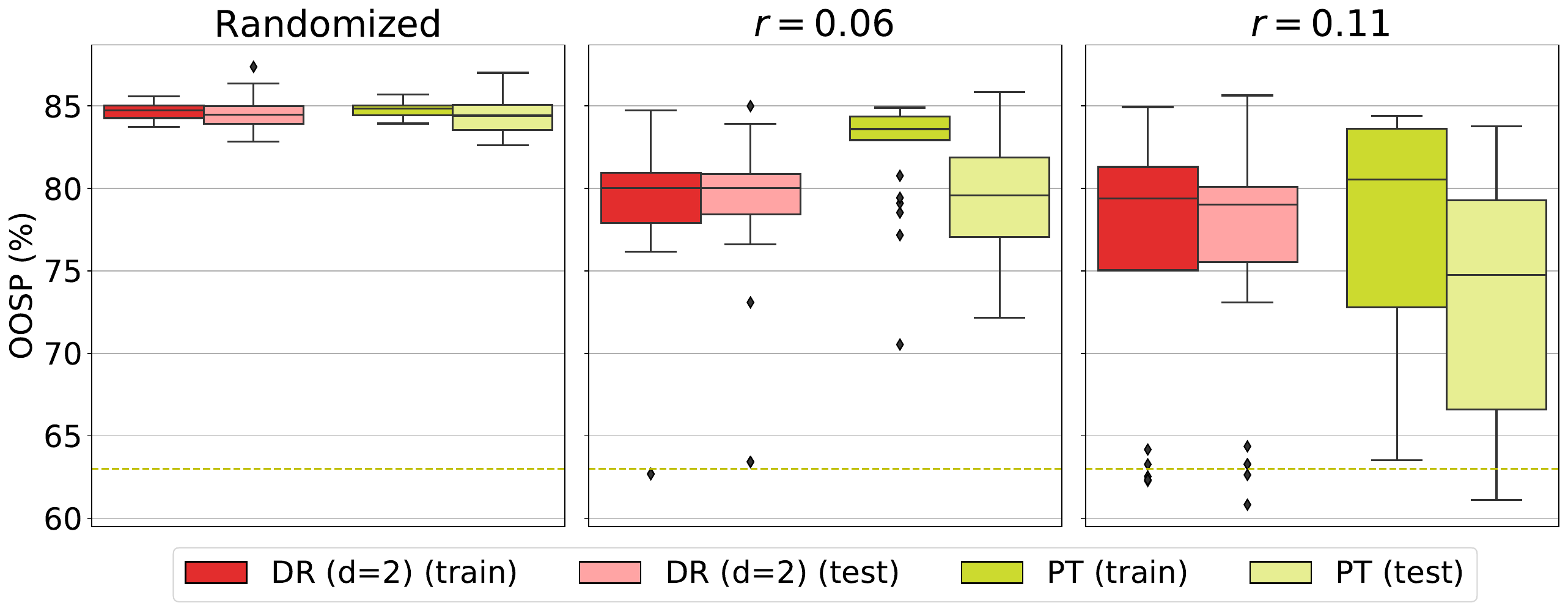}
\centering
\caption{\newnj{A comparison of the doubly robust method's performance to the formulation proposed by \cite{zhou2023offline} (PT), on the training and testing sets. Both methods are over trees of depth $d=2$.}}
\label{fig:warfarin_overfitting}
\end{figure}

\newnj{
\section{Policytree Overfitting}\label{sec:pt_overfitting}
In Figure~\ref{fig:warfarin_overfitting}, we illustrate that Policytree (PT) -- despite optimizing the same objective as our doubly robust method (DR) -- leads to worse out-of-sample performance because it learns from continuous features (as opposed to our method, which takes in discrete features). While PT's performance on the training set is more often than not better than DR, it does not generalize well on the testing set; in contrast, our method \revision{maintains similar performance between the training and testing sets.}}

\newnj{
\section{Scale of Proposed MIO}\label{sec:growth}
Table~\ref{tab:growth_mio} shows the growth of the number of constraints, continuous variables, and binary variables of formulations~\eqref{eq:robust},~\eqref{eq:ipw_mio}, and~\eqref{eq:dm_mio}. This growth is a function of the characteristics of the observational data (number of treatment options, size of covariates, etc.) and tree depth. \revision{It also assumes that we relax the integrality requirement on $z$ and $w$.}

\begin{table}[h]
    \centering
    \caption{\newnj{Summary of growth of the number of constraints and decision variables in formulations~\eqref{eq:robust},~\eqref{eq:ipw_mio}, and~\eqref{eq:dm_mio}.}}
    \begin{tabular}{lll}
    \# Constraints & \# Continuous vars. & \# Binary vars. \\
    \toprule
      $\sets{O}(2^d |\sets{K}||\sets{X}|)$ & $\sets{O}(2^{d+1} |\sets{K}||\sets{X}|)$& $\sets{O}(2^d |\sets{F}| \max_f\Theta(f))$  \\

    \end{tabular}
    \label{tab:growth_mio}
    \end{table}
}

\section{Proofs}
\label{section:proofs}

\proof{Proof of Proposition~\ref{prop:exact_IPW}.}
We introduce additional notation to help formalize our claim.
For any tree based policy $\pi \in \Pi_d$, define 
$$\QipwN(\pi) := \mathbb E\left[ \frac{ \mathbb I (K= \pi(X)) Y }{ \muhatN (K,X) } \right].$$
\newsa{To simplify the analysis, without loss of generality, we assume that the potential outcomes have been normalized, ensuring that $|Y(k)|\leq 1$ holds for all $k\in \mathcal{K}$.} We note that if all policies $\pi \in \Pi_d$ are optimal, the statements in the proposition follow immediately, and thus henceforth focus on the case where $\Pi_d \backslash \sets S^\star$ is not empty. 

We begin by showing that~$\vhatINipw \rightarrow v^\star$ w.p.1. Fix $\epsilon >0$ and a policy $\pi \in \Pi_d$. Since $\mu$ and $\muhatN$ are bounded away from $0$, $\exists m_0 >0$ and $N_0 \in \mathbb Z_+$ such that 
\begin{equation}
\mu (K,X) > m_0 \text{ and } \muhatN(K,X)>m_0 \quad \forall K \in \sets K, \;X \in \sets X, \text{ and } N \geq N_0.
\label{eq:lower_bound_m0}
\end{equation}
Define the function $g: (m_0,\infty) \rightarrow \mathbb R$ through $g(x):= \frac{1}{x}$. Since~$g$ is uniformly continuous on the interval $(m_0,\infty)$, equation~\eqref{eq:lower_bound_m0} implies that there exists $\delta_1>0$ such that 
\begin{align}
 \left|\muhatN (K,X) - \mu (K,X)\right|<\delta_1 \quad \implies \quad \left|\frac{1}{\muhatN (K,X)} - \frac{1}{\mu (K,X)}\right|  < \frac{\epsilon}{2} \quad \forall K \in \sets K, \;X \in \sets X.\label{eq:uniform_cont}
\end{align}
Moreover, since $\muhatN$ converges almost surely to $\mu$ \newsa{and the fact that the support set of $\mu$ is finite}, it follows that there exists \newsa{$N^{\pi}\geq N_0$} such that
\begin{align}
 \max_{X,K} \;\left|\muhatN (K,X) - \mu (K,X)\right|<\delta_1 \quad \text{w.p.1 } \quad \forall N\geq \newsa{N^{\pi}}.\label{eq:almost_sure_convergence}
\end{align}
Therefore, it follows from equations~\eqref{eq:uniform_cont} and~\eqref{eq:almost_sure_convergence} that
$$ \max_{X,K} \; \left|\frac{1}{\muhatN (K,X)} - \frac{1}{\mu (K,X)}\right|  < \frac{\epsilon}{2} \quad \text{w.p.1 } \quad \forall N\geq \newsa{N^{\pi}}.$$
We then have
\begin{equation*}
\renewcommand{\arraystretch}{1.3}
\begin{array}{ccl}
    \displaystyle \sup_{X,K,Y} \; \left|\frac{\mathbb I (K= \pi(X)) Y}{\muhatN (K,X)} - \frac{\mathbb I (K= \pi(X)) Y}{\mu (K,X)}\right|  & \leq &
    \displaystyle \newsa{\max_{X,K}} \; \left|\frac{\mathbb I (K= \pi(X)) }{\muhatN (K,X)} - \frac{\mathbb I (K= \pi(X))}{\mu (K,X)}\right|   \\
    & <  &
    \displaystyle \frac{\epsilon}{2} \quad \text{w.p.1 } \quad \forall N\geq \newsa{N^{\pi}},
\end{array}
\end{equation*}
%
%
where the \newsa{supremum and maximum} above are taken over $X \in \sets X$, $Y\in \sets Y$, $K\in \sets K$. The first inequality above follows since $|Y|\leq 1$ and implies that 
\begin{equation}
\renewcommand{\arraystretch}{2}
\begin{array}{cl}
    & \; \displaystyle \mathbb E\left( \frac{ \mathbb I (K= \pi(X)) Y }{ \mu (K,X) } - \frac{ \mathbb I (K= \pi(X)) Y }{ \muhatN (K,X) }\right) \\
     = &
    \displaystyle \sum_{x,k,y} \left( \frac{ \mathbb I (k= \pi(x)) y }{ \mu (k,x) } - \frac{ \mathbb I (k= \pi(x)) y }{ \muhatN (k,x) }\right) \mathbb P(X=x,K=k,Y=y) \\
    < & \displaystyle  \frac{\epsilon}{2} \quad \text{w.p.1 } \quad \forall N\geq \newsa{N^{\pi}}.
\end{array}
\end{equation}
%
%
\newsa{ Note that in the equation above, for the sake of clarity and ease of notation, we employ summation instead of integration}. By definition of $\Qipw$ and $\QipwN$ and from our conditional exchangeability assumption, see~\cite{hernan2019causal}, it then follows that 
\begin{equation}
    \left|\Qipw (\pi) - \QipwN (\pi)\right|= \left|Q (\pi) - \QipwN (\pi)\right|< \frac{\epsilon}{2} \quad \text{w.p.1 } \quad \forall N\geq \newsa{N^{\pi}}.
\label{eq:Qbound1}
\end{equation}
At the same time, by the strong law of large numbers, there exists \newsa{$I^{\pi}$} such that 
\begin{equation}
    \left|\QipwIN (\pi) - \QipwN (\pi)\right| < \frac{\epsilon}{2} \quad \text{w.p.1 } \quad \forall I\geq \newsa{I^{\pi}}.
\label{eq:Qbound2}
\end{equation}
Then, equations~\eqref{eq:Qbound1} and~\eqref{eq:Qbound2} imply that
\begin{equation}
    \left|\QipwIN (\pi) - Q (\pi)\right| \leq \epsilon \quad \text{w.p.1 } \quad \forall N\geq \newsa{N^{\pi}} \text{ and } I \geq \newsa{I^{\pi}}.
    \label{eq:Qbound2_pointwise}
\end{equation}

Since the set $\Pi_d$ of decision trees of depth at most~$d$ is finite, and the union of a finite number of sets each of measure zero also has measure zero, \newsa{the pointwise convergence given in equation~\eqref{eq:Qbound2_pointwise} also results in uniform convergence, meaning that, there exist $\displaystyle N_1 := \max_{\pi \in \Pi_d}N^{\pi}$ and~$\displaystyle I_0 := \max_{\pi \in \Pi_d}I^{\pi}$} such that
$$\max_{\pi \in \Pi_d}\left|\QipwIN (\pi) - Q (\pi)\right| \leq \epsilon \quad \text{w.p.1 }  \newsa{\quad \forall I \geq I_0 \text{ and } N \geq N_1.}$$
Since $|\vhatINipw - \vstar| \leq \epsilon$ \newsa{and the choice of $\epsilon$ was arbitrary}, then w.p.1 $\vhatINipw \rightarrow \vstar$ for all $I \geq I_0$ and $N \geq N_1$. This completes the first part of the proof.

We now show that $\shatINipw \subseteq \sets S^\star$ w.p.1 for $I$ and $N$ large enough. Consider the difference in objective value between the best and second best policies, given by
$$\rho := \max_{\pi \in \Pi_d \setminus \sstar} Q (\pi) - \vstar.$$
Since for any $\pi \in \Pi_d \setminus \sstar$ it holds that $Q (\pi) < \vstar$ and since the set $\Pi_d$ is finite, it follows that $\rho<0$.
Let $I$ and $N$ be large enough such that $\epsIN < - \frac{\rho}{2}$. Then, w.p.1, $\vhatINipw > \vstar + \frac{\rho}{2}$, and for any $\pi \in \Pi_d \setminus \sstar$ it holds that $\QipwIN (\pi) < \vstar+ \frac{\rho}{2}$. It follows that if $\pi \in \Pi_d \setminus \sstar$, then $\QipwIN < \vhatINipw$ and hence $\pi$ does not belong to set $\shatINipw$. As a result, for large enough $I$ and $N$, w.p.1 the inclusion $\shatINipw \subseteq \sstar$ holds, which completes the proof. \hfill \halmos
\endproof

We make use of the following auxiliary lemma in the proof of Proposition~\ref{prop:exact_DR}.

\begin{lemma}[\citet{bartle2001modern}]\label{lemma:basic_uniform_convergence}
Suppose $f_n: D \rightarrow \mathbb R$ and $g_n: D \rightarrow \mathbb R$ are sequences of functions which converge uniformly to $f,g: D \rightarrow \mathbb R$, respectively. Then, if $f$ and $g$ are bounded, i.e., $\exists B>0 \text{ such that } |f(x)|<B \text{ and } |g(x)|<B$, $f_ng_n$ converges uniformly to $fg$.
\end{lemma}

\proof{Proof of Proposition~\ref{prop:exact_DR}.} We introduce additional notation to help formalize our claim.
For any tree based policy $\pi \in \Pi_d$, define 
$$\QdrN(\pi) \; :=  \; \mathbb E\left[ \nuhatN_{\pi(X)}(X) + (Y-\nuhatN_{\pi(X)}(X))\frac{ \mathbb I (K= \pi(X)) }{ \muhatN(K,X) } \right].$$
Since $\hat \mu(K,X)$ and $\hat \nu_K(X)$ are bounded, there exists $B>0$ such that $|\hat \mu(K,X)|<B$ and $|\hat \nu_K(X)| < B \quad \forall X \in \sets X \text{ and } K \in \sets K$. Fix $\epsilon >0$ and a policy $\pi \in \Pi_d$. From the definition of $\hat \mu$, uniform continuity of the inverse function (bounded away from $0$), and the assumption that \newsa{$\sets X$ is finite and $\sets Y$ is bounded, $\exists N^{\pi} \in \mathbb Z_+$ such that}
\begin{equation}\label{eq:prop_exact_DR_proof_1}
\renewcommand{\arraystretch}{1.3}
\begin{array}{ccl}
    \displaystyle \sup_{X,K,Y} \; \left|\frac{\mathbb I (K= \pi(X)) Y}{\muhatN (X,K)} - \frac{\mathbb I (K= \pi(X)) Y}{\hat \mu (X,K)}\right| 
    & < &
    \; \displaystyle \newsa{\max_{X,K}} \; \left|\frac{\mathbb I (K= \pi(X)) }{\muhatN (X,K)} - \frac{\mathbb I (K= \pi(X)) }{\hat \mu (K,X)}\right| \\
     & < & 
     \displaystyle  \frac{\epsilon}{4(B+1)} \quad \forall N \geq \newsa{N^{\pi}}.
\end{array}
\end{equation}
Also, by definition of \newsa{$\hat \nu$, $\exists N_0 \in \mathbb Z_+$} such that
\begin{equation}\label{eq:prop_exact_DR_proof_2}
\max_{X,K} \; \left|\nuhatN_{K}(X)- \hat \nu_{K}(X)\right| < \frac{\epsilon}{4(B+1)} \quad \forall N \geq N_0.
\end{equation}
Thus, using~\eqref{eq:prop_exact_DR_proof_1},~\eqref{eq:prop_exact_DR_proof_2}, and the assumption that $\hat \nu$ and $\hat \mu(K,X)$ are bounded, by Lemma~\ref{lemma:basic_uniform_convergence}, \newsa{$\exists N^{\pi}_1 \geq N^{\pi},N_0$ such that}
\begin{equation}\label{eq:prop_exact_DR_proof_3}
\max_{X,K} \; \left|\frac{\mathbb I (K= \pi(X)) \nuhatN_{K}(X) }{\muhatN (K,X)} - \frac{\mathbb I (K= \pi(X)) \hat \nu_{K}(X) }{\hat \mu (K,X)}\right| < \frac{2\epsilon B}{4(B+1)} \quad \forall N \geq \newsa{N^{\pi}_1}.
\end{equation}
Moreover, from~\eqref{eq:prop_exact_DR_proof_1}, \eqref{eq:prop_exact_DR_proof_2}, and~\eqref{eq:prop_exact_DR_proof_3}, we can conclude that for $N \geq \newsa{N^{\pi}_1}$,
$$\sup_{X,K,Y} \left|\left\{\nuhatN_{\pi(X)}(X) + (Y-\nuhatN_{\pi(X)}(X))\frac{ \mathbb I (K= \pi(X)) }{ \muhatN(K,X) }\right\} - 
\left\{\hat \nu_{\pi(X)}(X) + (Y-\hat \nu_{\pi(X)}(X))\frac{ \mathbb I (K= \pi(X)) }{ \hat \mu(K,X) }\right\}\right| < \frac{\epsilon}{2}.$$
Similar to the proof of Proposition~\ref{prop:exact_IPW}, for a given policy $\pi \in \Pi_d$, for $N \geq \newsa{N^{\pi}_1}$,
$$\left|\Qdr (\pi) - \QdrN (\pi)\right|< \frac{\epsilon}{2}.$$
%

Finally, we know that if w.p.1, either $\hat \mu(k,x) = \mu(k,x)$ or $\hat \nu_k(x) = \nu_k(x)$, for all $x \in \sets X$ and $k\in \sets K$, then $Q(\pi) = \Qdr (\pi)$, see for example~\cite{lunceford2004stratification}. Since the assumption in the previous clause is satisfied w.p.1 (as a result of the almost sure convergence of either $\muhatN$ or $\nuhatN_K$), we conclude that, 
$$\left|Q (\pi) - \QdrN (\pi)\right|< \frac{\epsilon}{2} \quad \text{w.p.1} \quad \forall N\geq \newsa{N^{\pi}_1.}$$
The rest of the proof is omitted as it can be derived by following the same logic as in the proof of Proposition~\ref{prop:exact_IPW}.\hfill \halmos
\endproof

\proof{Proof of Proposition~\ref{prop:exact_DM}.}
We omit the proof due to its similarity to the proof of Proposition~\ref{prop:exact_DR}.\flushright \halmos
\endproof

\end{document}